\documentclass[11pt]{article}

\usepackage[final]{acl}

\usepackage{times}
\usepackage{latexsym}

\usepackage[T1]{fontenc}

\usepackage[utf8]{inputenc}

\usepackage{microtype}

\usepackage{inconsolata}

\usepackage{graphicx}

\usepackage{xspace}
\usepackage{booktabs}
\usepackage{amsmath}
\usepackage{amsfonts}
\usepackage{amssymb}
\usepackage{array}
\usepackage{multirow}
\usepackage{enumitem}
\usepackage{longtable}
\usepackage{algorithm}
\usepackage[noend]{algpseudocode}
\usepackage[table]{xcolor}

\newcommand{\lct}{\textsc{LCT}\xspace}
\newcommand{\bpe}{\textsc{BPE}\xspace}

\newcolumntype{L}[1]{>{\raggedright\arraybackslash}p{#1}}
\title{Latent Core Tokenizer: Compress, but Meaningfully}

\author{ Felermino D. M. A. Ali\thanks{\hspace{0.1cm} Equal Contribution.}$^1$, Millicent Ochieng\footnotemark[1]$^1$, Ogbemi Ekwejunor-Etchie$^2$,\\ 
\textbf{Ade Famoti$^2$}, \textbf{Jacki O’Neill$^1$}, \textbf{Debjit Paul$^3$} \\
\footnotesize
$^1$Microsoft Research Africa, 
$^2$Microsoft Research Accelerator, 
$^3$Microsoft Research India
\\
}

\begin{document}
\maketitle


\begin{abstract}

Tokenizers are commonly optimized for compression, but a compact vocabulary does not necessarily distribute its capacity evenly across languages. We introduce the Latent Core Tokenizer (LCT), an approach that separates structural discovery from vocabulary construction. LCT uses Minimum Description Length, entropy-based boundary signals, and morphotactic constraints to identify reusable linguistic units before constructing a shared vocabulary. Across 104 languages with a 200K-token vocabulary, LCT achieves lower fertility and higher MorphScore than BPE, Unigram, and parity-aware BPE, while maintaining comparable cross-lingual disparity in tokenization cost. Across four multilingual downstream benchmarks, LCT improves aggregate score by 1.48, 1.83, and 2.00 points over BPE, Unigram, and parity-aware BPE, respectively. Our findings show that compression alone does not predict representation quality and highlight the importance of morphology-driven structural discovery and how frequency is used to allocate the final vocabulary across languages.

\end{abstract}

\section{Introduction}
\label{sec:introduction}

Tokenization determines the units through which language models represent text, with direct consequences for sequence length, vocabulary capacity, and the information available to the model. Subword methods such as Byte Pair Encoding (BPE) and Unigram tokenization make open vocabulary modeling practical by learning reusable units from corpus statistics \citep{sennrich-etal-2016-neural,kudo-2018-subword}. In multilingual settings, however, a shared vocabulary must serve languages that differ substantially in corpus size, script, and morphological structure. Frequency driven vocabulary learning can therefore favor patterns from high resource languages and lead to uneven tokenization efficiency across languages \citep{rust-etal-2021-good,foroutan-etal-2026-parity}. At the same time, work on morphology aware tokenization shows that the units that best preserve linguistic structure need not be the same units that optimize compression \citep{vemula-etal-2025-rethinking,asgari-etal-2026-morphbpe}. Conventional subword tokenizers typically use the same vocabulary learning process both to identify reusable units and to determine the final tokens exposed to the model.

\begin{figure}
    \centering
    \includegraphics[width=1\linewidth]{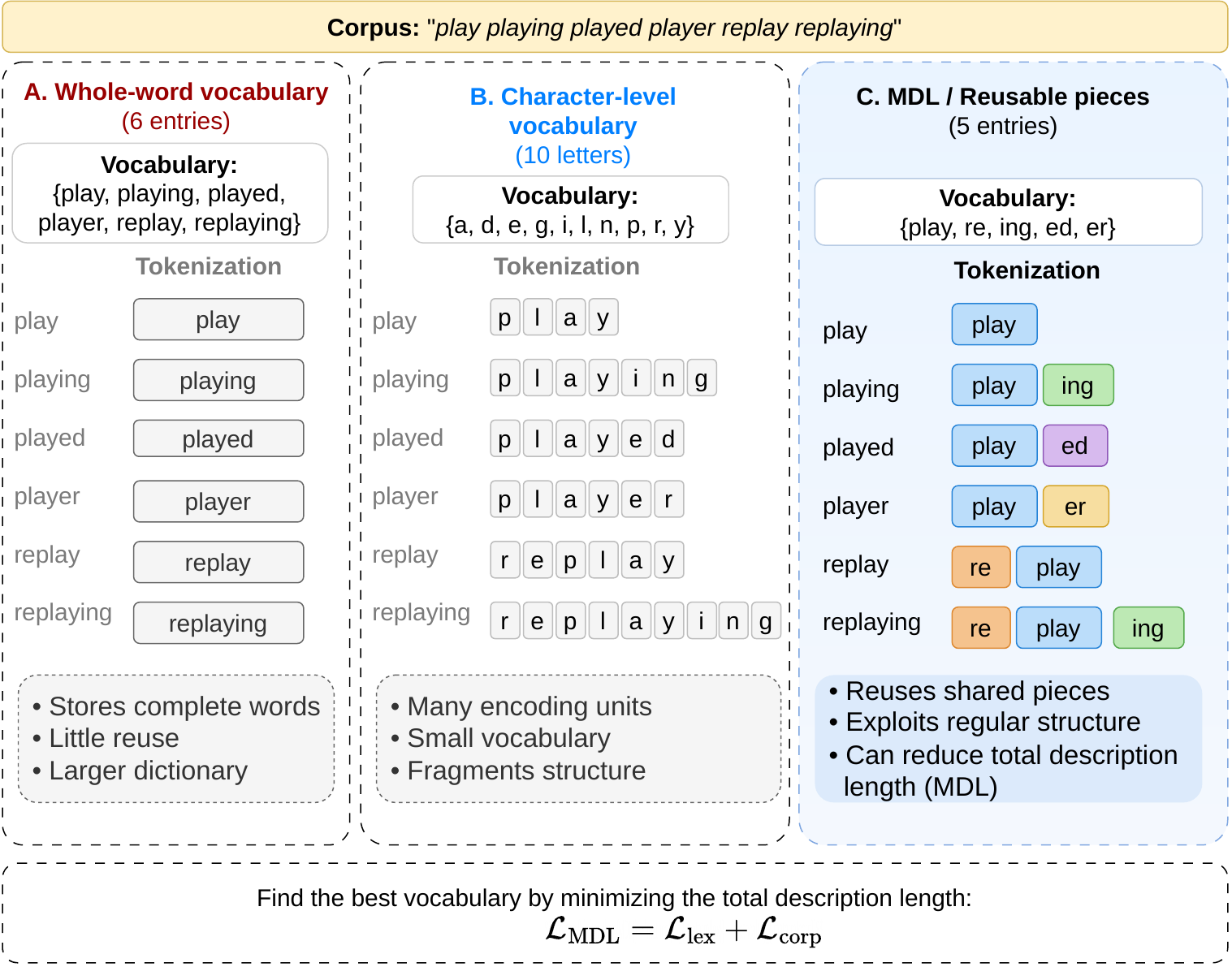}
  \caption{Illustration of minimum description length (MDL)-based tokenization compared with whole-word and character-level vocabularies. By identifying reusable subword pieces that capture shared patterns across word forms, an MDL-based tokenizer balances vocabulary and corpus encoding efficiency, minimizing the total description length $\mathcal{L}_{\mathrm{MDL}}$, where $\mathcal{L}_{\mathrm{lex}}$ denotes the cost of encoding the vocabulary and $\mathcal{L}_{\mathrm{corp}}$ denotes the cost of encoding the corpus using that vocabulary.}
\label{fig:MDL}
\end{figure}

We introduce the Latent Core Tokenizer (LCT), which separates latent structural discovery from surface vocabulary construction. LCT first learns reusable internal structure from multilingual text using minimum description length segmentation, branching entropy, and morphotactic refinement under language-balanced evidence. These latent pieces guide vocabulary learning but need not correspond to the final tokens consumed by the model. LCT then constructs a fixed surface vocabulary by selectively recombining latent pieces according to their utility for efficient representation. This allows structural information to guide vocabulary construction without constraining the final surface segmentation.

We evaluate LCT across 104 languages against BPE, Unigram, and Parity-Aware BPE under matched corpus, vocabulary budget, and encoder architecture. LCT achieves lower fertility and higher MorphScore than all three baselines while maintaining cross-lingual disparity in tokenization cost comparable to BPE. Under controlled encoder pretraining, LCT improves the aggregate downstream score by 1.48, 1.83, and 2.00 points over BPE, Unigram, and Parity-Aware BPE, respectively.

Our contributions are threefold: (i) we introduce LCT, a tokenizer design that decouples latent structural discovery from surface vocabulary construction; (ii) we show across 104 languages that this design jointly improves tokenization efficiency and morphological alignment while maintaining comparable cross-lingual tokenization-cost disparity; and (iii) we show through controlled multilingual encoder pretraining that these improvements extend to downstream representations. To support reproducibility and further research, we will release the LCT implementation, trained tokenizer configurations and vocabularies, and evaluation code upon publication.


\section{Latent Core Tokenizer}
\label{sec:LCT}

The central idea of \lct is that the units used to \emph{understand} a word during tokenizer training need not be the units emitted to a neural model. Conventional frequency-based tokenizers usually discover reusable pieces and compress text in the same operation. \lct separates these goals: it first learns a latent structural analysis, then uses that analysis to construct a fixed surface vocabulary. The latent pieces guide vocabulary learning but are not labels or additional inputs presented to the encoder.

Given language-indexed corpora $\{\mathcal{D}_{\ell}\}$ and vocabulary budget $K$, training has four steps:

\begin{enumerate}[leftmargin=*]
  \item \textbf{Balance the evidence.} Word frequencies are rescaled so that each language contributes approximately equal total mass. This is not an equal vocabulary quota: a piece reused across languages can still receive more support.

  \item \textbf{Discover reusable structure.} A minimum-description-length segmenter trades the cost of storing pieces against the savings from reusing them (see Figure~\ref{fig:MDL}). Branching entropy proposes likely internal boundaries, and a morphotactic model distinguishes prefix-, core-, and suffix-like behavior. These signals refine an internal analysis without requiring gold morphological annotations.

  \item \textbf{Construct surface tokens.} Supported characters and latent pieces seed the vocabulary. A ranked packer repeatedly merges the adjacent pair that saves the most weighted token occurrences, allowing frequent combinations and complete words to become single tokens. Packing continues until the exact learned-vocabulary budget is filled.

  \item \textbf{Guarantee usable encoding.} The learned surface vocabulary is assigned fixed token IDs. At runtime, minimum-token dynamic programming selects among known surface pieces, while UTF-8 byte tokens provide exact reversible coverage for any unseen string.
\end{enumerate}

This separation is the underlying principle: structural discovery decides \emph{which building blocks are worth preserving}, whereas surface packing decides \emph{which combinations are worth an embedding slot}. A productive stem can therefore influence vocabulary construction even when a frequent whole word containing it is ultimately emitted as one token. The remainder follows these stages in the same order, introducing each mathematical expression where it is needed. Algorithm~\ref{alg:LCT-training} then collects the complete procedure in one place.

\label{sec:algorithm-mathematics}

We now follow the four stages in detail, beginning with the evidence used to learn the latent analysis.

\paragraph{Balanced evidence.}

For language $\ell$, Algorithm~\ref{alg:LCT-training} first constructs a word-frequency table $F_\ell$ with total mass $N_\ell=\sum_w F_\ell(w)$. Its contribution is multiplied by $N_{\max}/N_\ell$, so its effective total is approximately $N_{\max}$ regardless of corpus size. The rounding floor of one prevents an observed word type from disappearing. Summing the rescaled tables gives $\widetilde{F}$, which is used consistently by all subsequent training stages. Thus, ``balanced'' means equal aggregate frequency mass, not an equal quota of final vocabulary entries.

\paragraph{Minimum description length (MDL) segmentation.}

Let $\Sigma$ be the observed character alphabet, let $\mathcal{M}$ be the current latent-piece lexicon, let $c(m)$ be the frequency of piece $m$ under the current segmentations, and let $C=\sum_m c(m)$. The implemented Morfessor-style objective can be written

\begin{align}
\mathcal{L}_{\mathrm{MDL}}
  &= \mathcal{L}_{\mathrm{lex}}+\mathcal{L}_{\mathrm{corp}},\\
\mathcal{L}_{\mathrm{lex}}
  &= C_\Sigma\sum_{m\in\mathcal{M}}(|m|+1),\\
\mathcal{L}_{\mathrm{corp}}
  &= C\log_2 C-\sum_{m\in\mathcal{M}}c(m)\log_2 c(m),\\
C_\Sigma
  &= \lambda\max\left(1,\log_2(|\Sigma|+1)\right),
\label{eq:mdl-expanded}
\end{align}

where $\lambda=1.5$ is the lexicon weight. The corpus term is the unigram code length $\sum_m -c(m)\log_2[c(m)/C]$. A whole word costs one corpus symbol but must be stored as a long lexicon entry; character segmentation minimizes lexicon storage but creates many corpus symbols. A reusable stem or affix is favored when its repeated corpus savings exceed its one-time lexicon cost.

Training begins with whole words. For each weighted word, its current analysis is temporarily removed from the counts and a dynamic program finds the minimum incremental description-length segmentation with pieces of at most 20 characters. Counts are then restored using the selected analysis. This coordinate-style resegmentation repeats for at most eight epochs or until no analysis changes.

\paragraph{Branching entropy.}

MDL alone can miss a productive boundary when several continuations occur after the same context. For a left context $x$, the forward branching entropy is

\begin{equation}
H_{\rightarrow}(x)=
-\sum_{a\in\Sigma\cup\{\mathrm{END}\}}
p(a\mid x)\log_2 p(a\mid x).
\end{equation}

A backward model applies the same calculation to reversed text. At internal position $i$ of word $w$, the normalized bidirectional score is

\begin{equation}
b_i(w)=\frac{1}{2}\left[
\frac{H_{\rightarrow}(w_{<i})}{\log_2(|\Sigma|+1)}
+
\frac{H_{\leftarrow}(\operatorname{rev}(w_{\ge i}))}
     {\log_2(|\Sigma|+1)}
\right].
\label{eq:branching-score}
\end{equation}

Positions with $b_i(w)\ge q$ become hard boundaries: the segmentation dynamic program may end or begin a piece there but may not place a multi-character piece across the boundary.

The threshold is not selected with morphological labels. We evaluate
$q\in\{0,0.28,0.34,0.40,0.46,0.52,0.58\}$.
For each value, a segmenter is trained on four of every five sorted word types. The chosen value minimizes

\begin{equation}
q^\star=\arg\min_q\left[
\mathcal{L}_{\mathrm{lex}}(q)
+
\sum_{w\in\mathcal{H}}\widetilde{F}(w)
\mathcal{C}_q(w)
\right],
\label{eq:entropy-selection}
\end{equation}

where $\mathcal{H}$ is the held-out type set and $\mathcal{C}_q(w)$ is the unigram code length of the segmentation produced by the model trained with threshold $q$.

\paragraph{Morphotactic refinement.}

The resulting pieces are assigned latent categories
$\mathcal{Z}=\{\mathrm{PRE},\mathrm{STM},\mathrm{SUF},\mathrm{NON}\}$, corresponding to prefixes, stems, suffixes, and non-affix units, respectively.
Each piece emits a feature tuple containing its length bucket, left-context entropy bucket, right-context entropy bucket, and whether it occurs as a standalone word. A first-order hidden Markov model selects
\begin{equation}
\begin{aligned}
z^\star
= \arg\max_{z_1,\ldots,z_r}
\Bigg[
&\sum_{i=1}^{r}
    \log P\!\left(f(m_i)\mid z_i\right) \\
&+
\sum_{i=1}^{r+1}
    \log P\!\left(z_i\mid z_{i-1}\right)
\Bigg].
\end{aligned}
\label{eq:morphotactics}
\end{equation}


under a grammar approximating $\mathrm{PRE}^{*}\,\mathrm{STM}\,(\mathrm{STM}\mid\mathrm{SUF})^{*}$, with \textsc{NON} allowed as a noise state. Transition and emission tables use additive smoothing and are re-estimated for five hard-EM iterations using Viterbi assignments. \textsc{NON} fragments are merged into a neighbor, removing low-confidence boundaries without changing the surface string. Corpus-supported core and affix merges then reduce residual fragmentation while preserving the latent decomposition.

\paragraph{Surface packing.}

The latent analyzer produces a weighted multiset $C$ of piece sequences. For an adjacent pair $(a,b)$, the packer maintains the support score,
$g(a,b)=\sum_{z} C(z)\,n_z(a,b)$,
where $n_z(a,b)$ counts current adjacent occurrences in sequence $z$. The highest-score pair is replaced in all affected sequences, its concatenation $ab$ is added to the fixed vocabulary, and affected pair scores are updated. Because each replacement normally removes one emitted symbol, $g$ estimates the frequency-weighted token saving. Deterministic lexical tie-breaking and recorded merge ranks make the fitted vocabulary reproducible.

\section{Experimental Setup}
\label{sec:experimental-setup}
We evaluate LCT under controlled conditions using the same multilingual corpus, vocabulary budget, and encoder architecture across all tokenizer variants.

\subsection{Multilingual Corpus}
\label{sec:corpus}

All tokenizer and encoder variants are trained on the same multilingual corpus constructed from FineWeb and FineWeb2. English is sourced from FineWeb, while the remaining 103 languages come from filtered FineWeb2 subsets, covering high-, medium-, and low-resource settings. For a controlled comparison, document selection is deterministic, duplicates are removed, and the same training and validation documents are used across all configurations. We use a deterministic 1\% validation split.
We use all datasets and artifacts in accordance with their respective licenses and terms of use.

\paragraph{Sampling}

To reduce the dominance of high-resource languages while preserving the natural distribution of the data available across languages, we apply temperature-based language sampling:
$p(l)=\frac{n(l)^{0.3}}{\sum_j n(j)^{0.3}}$,
where $n(l)$ is the available corpus size for language $l$ in UTF-8 bytes. We then allocate language-specific quotas from a shared budget of 8~GiB for training and approximately 82~MiB for validation. Documents are sampled deterministically without truncation whenever possible, with oversampling recorded for languages with insufficient data. This procedure keeps the corpus and data distribution fixed across tokenizer arms, allowing downstream differences to be attributed to tokenization rather than corpus composition. Appendix~\ref{app:data} shows the resulting language distribution, statistics, and sampling allocations.

\subsection{Evaluation}
\label{sec:evaluation}

We compare LCT with BPE, Unigram, and Parity-Aware BPE using both intrinsic tokenizer metrics and downstream encoder evaluation.

\paragraph{Intrinsic evaluation.}
\begin{table*}[h]
  \centering
  \footnotesize
  \setlength{\tabcolsep}{5pt}
  \begin{tabular}{@{}lccccccc@{}}
    \toprule
     & Comp. Rate $\uparrow$
     & TTR $\uparrow$
     & Vocab. Util. $\uparrow$
     & Fertility $\downarrow$
     & R\'enyi ($\alpha{=}2.5$) $\uparrow$
     & MorphScore F1 $\uparrow$
     & Gini $\downarrow$ \\
    \midrule
    \texttt{LCT-MEM}
      & \textbf{4.866}
      & \textbf{0.1781}
      & \textbf{0.2526}
      & \textbf{1.999}
      & 0.4472
      & \textbf{0.5209}
      & \textbf{0.1341} \\
    BPE
      & 4.295
      & 0.1349
      & 0.2168
      & 2.241
      & 0.4613
      & 0.4410
      & 0.1342 \\
    Parity-Aware BPE
      & 3.340
      & 0.0630
      & 0.1301
      & 2.742
      & \textbf{0.4879}
      & 0.3186
      & 0.1563 \\
    Unigram
      & 3.195
      & 0.0878
      & 0.1898
      & 2.920
      & 0.3142
      & 0.3109
      & 0.1435 \\
    \bottomrule
  \end{tabular}

  \caption{Intrinsic tokenizer quality at 200k vocabulary.
  \texttt{LCT-MEM} denotes the LCT configuration combining MDL, branching entropy, and morphotactic refinement. Per-language results are displayed in
  Table~\ref{tab:tokeval-lang-quality} and
  Table~\ref{tab:tokeval-lang-efficiency}.}
  \label{tab:intrinsic-expanded}
\end{table*}

We evaluate reconstruction, efficiency, morphological alignment, and multilingual vocabulary allocation. We follow a similar protocol to \citet{foroutan-etal-2026-parity}, measuring compression rate, type-token ratio (TTR), vocabulary utilization, fertility, MorphScore F1, and Gini. \textbf{Fertility} is the mean number of tokens per word. \textbf{Vocabulary Utilization} is the share of the 200k vocabulary the corpus activates. Fertility and vocabulary utilization are measured on UDHR translations\footnote{\url{https://huggingface.co/datasets/community-datasets/udhr}} over 93 languages. \textbf{MorphScore} \citep{arnett2025alignment} is evaluated on languages available in the dataset with at least 100 eligible examples and reports the official frequency-weighted F1, for a total of 65 languages with sufficient gold items. \textbf{Compression rate} is UTF-8 bytes per token. \textbf{TTR} is the type-token ratio, defined as the number of distinct tokens emitted divided by the total number of tokens emitted across the corpus. \textbf{R\'enyi} efficiency at $\alpha{=}2.5$ rewards a flatter token-frequency distribution. \textbf{Gini} is the Gini Token Inequality Coefficient over per-byte cost, using TokEval's\footnote{\url{https://github.com/cimeister/tokenizer-intrinsic-evals}} default normalisation.

\paragraph{Extrinsic evaluation.}

We evaluate the pretrained encoder representations on four benchmarks: XNLI \citep{conneau-etal-2018-xnli}, Belebele \citep{bandarkar-etal-2024-belebele}, XStoryCloze \citep{mihaylov-frank-2017-story}, and PAWS-X \citep{yang-etal-2019-paws}. For each tokenizer, we train one DeBERTa-v3-style XSmall encoder (see configuration in Table~\ref{tab:encoder-settings}) using pretraining seed 42, keeping the architecture and training budget identical across variants. We then fine-tune each pretrained encoder using five seeds (11, 29, 47, 53, and 71), with the learning rate selected separately for each tokenizer on the development set. This paired design allows differences across fine-tuning runs to be attributed to the tokenizer while holding pretraining fixed. Additional extrinsic evaluation details are provided in Appendix~\ref{app:settings}.

\section{Results}
\label{sec:results}

We first report intrinsic tokenizer performance and component-level ablations, followed by downstream evaluation of the resulting multilingual encoder representations.

\subsection{Intrinsic Evaluation}
\label{sec:intrinsic-results}

\begin{figure*}
    \centering
    \includegraphics[width=0.9\linewidth]{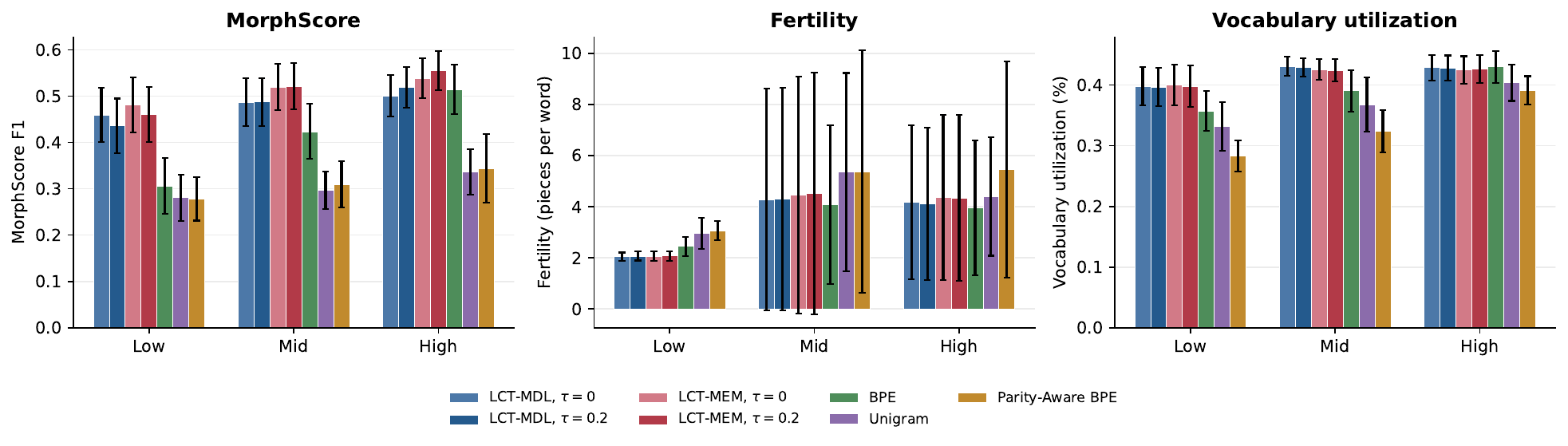}
    \caption{Intrinsic tokenizer performance across low-, mid-, and high-resource language tertiles. Bars show macro-averages and error bars indicate 95\% confidence intervals across languages. MorphScore uses 65 languages, while fertility and vocabulary utilization use the 93 languages covered by TokEval. All configurations use a 200k vocabulary.}
    \label{fig:resource-intrinsic}
\end{figure*}

Table~\ref{tab:intrinsic-expanded} compares \texttt{LCT-MEM} with BPE, Parity-Aware BPE, and Unigram. \texttt{LCT-MEM} has the highest compression rate (4.866 bytes per token), TTR (0.1781), and vocabulary utilization (25.26\%), and the lowest fertility (1.999 tokens per word). It also obtains the highest MorphScore F1 (0.5209), compared with 0.4410 for BPE, 0.3186 for Parity-Aware BPE, and 0.3109 for Unigram. Its Gini (0.1341) is effectively tied with BPE (0.1342) and lower than Parity-Aware BPE (0.1563) and Unigram (0.1435). Parity-Aware BPE is stronger only on Rényi efficiency. 

These results show that \texttt{LCT-MEM} provides the strongest balance of compression, vocabulary use, morphological alignment, and cross-language tokenization-cost equality among the compared systems. They establish the intrinsic behavior of the final tokenizer, but do not identify which design choices drive the gains. The ablation study in Section~\ref{sec:ablations} therefore separates the effects of language balancing, latent structural discovery, entropy, and morphotactic refinement.

\subsubsection{Ablation Study}
\label{sec:ablations}

Keeping the training corpus and round-trip coverage fixed, we vary language-frequency estimation, vocabulary size, latent structural discovery, and morphotactic induction.

\begin{table}[h]
  \centering
  \small
  \setlength{\tabcolsep}{3pt}

  \resizebox{\columnwidth}{!}{%
  \begin{tabular}{@{}rr l r r r r r r@{}}
    \toprule
    \multicolumn{2}{c}{Rank} & &
    \multicolumn{2}{c}{Fertility $\downarrow$} &
    \multicolumn{2}{c}{MorphScore $\uparrow$} &
    \multicolumn{2}{c}{Gini $\downarrow$} \\
    \cmidrule(lr){1-2}
    \cmidrule(lr){4-5}
    \cmidrule(lr){6-7}
    \cmidrule(lr){8-9}

    \#$_{\text{200k}}$ & \#$_{\text{256k}}$ &
    System &
    200k & 256k &
    200k & 256k &
    200k & 256k \\
    \midrule

    1 & 1 & \texttt{LCT-MEM} $\tau{=}0.2$ &
    2.023 & 1.935 & \textbf{0.5249} & \textbf{0.5556} & 0.1369 & 0.1409 \\

    2 & 2 & \texttt{LCT-MEM} $\tau{=}0$ &
    2.017 & 1.929 & 0.5209 & 0.5507 & \textbf{0.1341} & 0.1384 \\

    3 & 3 & \texttt{LCT-MEM} $\tau{=}0.4$ &
    2.047 & 1.960 & 0.5244 & 0.5543 & 0.1374 & 0.1422 \\

    4 & 4 & \texttt{LCT-ME} $\tau{=}0.8$ &
    2.053 & 1.963 & 0.5233 & 0.5549 & 0.1536 & 0.1570 \\

    5 & 8 & \texttt{LCT-MDL} $\tau{=}0$ &
    \textbf{1.962} & \textbf{1.876} & 0.4880 & 0.5245 & 0.1402 & 0.1439 \\

    6 & 6 & \texttt{LCT-MDL} $\tau{=}0.2$ &
    1.966 & 1.879 & 0.4916 & 0.5287 & 0.1423 & 0.1463 \\

    7 & 5 & \texttt{LCT-MDL} $\tau{=}0.4$ &
    1.973 & 1.890 & 0.4923 & 0.5295 & 0.1452 & 0.1492 \\

    8 & 7 & \texttt{LCT-MEM} $\tau{=}0.6$ &
    2.065 & 1.978 & 0.5188 & 0.5492 & 0.1418 & 0.1455 \\

    9 & 10 & \texttt{LCT-ME} $\tau{=}1$ &
    2.091 & 2.000 & 0.5169 & 0.5467 & 0.1551 & 0.1587 \\

    10 & 9 & \texttt{LCT-MDL} $\tau{=}0.6$ &
    1.992 & 1.907 & 0.4876 & 0.5261 & 0.1476 & 0.1517 \\

    \midrule

    25 & 25 & BPE &
    2.229 & 2.166 & 0.4410 & 0.4697 & 0.1342 & \textbf{0.1369} \\

    26 & 26 & Parity-Aware BPE &
    2.844 & 2.864 & 0.3186 & 0.3286 & 0.1563 & 0.1363 \\

    27 & 27 & Unigram &
    2.977 & 2.975 & 0.3109 & 0.3101 & 0.1435 & 0.1435 \\

    \bottomrule
  \end{tabular}%
  }

  \caption{Top 10 tokenizer configurations at 200k and 256k vocabulary sizes.
  Additional intrinsic evaluation results are provided in
  Figure~\ref{fig:intrinsic_summary}.}
  \label{tab:encoder-tokenizer-ranking}
\end{table}

\paragraph{Language-frequency estimation.}

We generalize Equation~\ref{eq:balance-temperature} with a temperature exponent $\tau$:
\begin{equation}
\widetilde{F}^{(\tau)}_{\ell}(w) =
F_{\ell}(w)
\left(\frac{N_{\max}}{N_{\ell}}\right)^{1-\tau}.
\label{eq:balance-temperature}
\end{equation}

Here $\tau=0$ recovers equal-total language balancing, while $\tau=1$ removes balancing and recovers raw relative language frequencies. Intermediate values partially balance languages by increasing the influence of smaller languages without forcing equal effective totals. Intermediate values partially balance languages by increasing the influence of languages with less available training data without forcing equal effective totals. We evaluate $\tau\in\{0.0,0.2,0.4,0.6,0.8,1\}$ to test equal, partial, and raw-frequency weighting.

\paragraph{Latent discovery and Morphotactics}

We vary latent discovery and morphotactic induction in a $2\times2$ design to isolate their individual and combined effects. The four variants are \texttt{LCT-MDL}, which uses MDL alone; \texttt{LCT-ME}, which adds branching entropy; \texttt{LCT-MM}, which adds morphotactic induction; and \texttt{LCT-MEM}, which combines MDL, entropy, and morphotactics. Every variant is trained at each value of $\tau$.

\paragraph{Vocabulary budget}

We repeat the full design at 200k and 256k learned entries to test whether the observed component effects persist across vocabulary sizes.

\begin{table*}[h]
\centering
\scriptsize
\setlength{\tabcolsep}{3pt}

\begin{tabular}{lccccrrrrrrr}
\toprule
& \multicolumn{4}{c}{Benchmark}
& \multicolumn{7}{c}{Aggregate accuracy (\%)} \\
\cmidrule(lr){2-5}
\cmidrule(lr){6-12}

Language
& XNLI
& Belebele
& XStoryCloze
& PAWS-X
& \shortstack{\texttt{LCT-MDL}\\$\tau=0$}
& \shortstack{\texttt{LCT-MDL}\\$\tau=0.2$}
& \shortstack{\texttt{LCT-MEM}\\$\tau=0$}
& \shortstack{\texttt{LCT-MEM}\\$\tau=0.2$}
& BPE
& Unigram
& \shortstack{Parity-Aware\\BPE} \\
\midrule

Arabic & $\checkmark$ & $\checkmark$ & $\checkmark$ & -- & 42.14 & \textbf{42.65} & 41.90 & 42.10 & 42.46 & 41.92 & 41.79\\
Basque & -- & $\checkmark$ & $\checkmark$ & -- & 42.71 & 42.83 & 42.78 & \textbf{43.10} & 42.76 & 41.34 & 42.78\\
Bulgarian & $\checkmark$ & $\checkmark$ & -- & -- & 39.27 & \textbf{40.25} & 39.68 & 40.18 & 38.98 & 37.84 & 38.99\\
Chinese & $\checkmark$ & $\checkmark$ & $\checkmark$ & $\checkmark$ & 49.89 & 50.04 & 50.11 & \textbf{51.40} & 47.30 & 47.77 & 45.99\\
English & $\checkmark$ & $\checkmark$ & $\checkmark$ & $\checkmark$ & 62.20 & 62.33 & 61.63 & \textbf{62.39} & 61.70 & 61.45 & 61.38\\
French & $\checkmark$ & $\checkmark$ & -- & $\checkmark$ & 53.30 & 53.65 & 53.44 & 53.54 & \textbf{53.74} & 51.60 & 50.82\\
German & $\checkmark$ & $\checkmark$ & -- & $\checkmark$ & 51.95 & 53.35 & 53.34 & \textbf{53.51} & 52.89 & 50.92 & 49.33\\
Greek & $\checkmark$ & $\checkmark$ & -- & -- & 38.05 & \textbf{39.03} & 36.90 & 38.06 & 37.84 & 36.10 & 37.57\\
Hindi & $\checkmark$ & $\checkmark$ & $\checkmark$ & -- & 43.00 & 42.27 & 42.39 & \textbf{43.56} & 40.13 & 39.99 & 39.53\\
Indonesian & -- & $\checkmark$ & $\checkmark$ & -- & \textbf{43.72} & 43.39 & 42.77 & 43.71 & 43.29 & 41.88 & 43.04\\
Japanese & -- & $\checkmark$ & -- & $\checkmark$ & 47.64 & 47.81 & 47.02 & \textbf{48.28} & 43.36 & 46.00 & 43.10\\
Korean & -- & $\checkmark$ & -- & $\checkmark$ & 47.60 & 49.01 & 47.66 & \textbf{49.97} & 46.62 & 46.71 & 46.50\\
Russian & $\checkmark$ & $\checkmark$ & $\checkmark$ & -- & 43.87 & 44.95 & 44.07 & 44.44 & \textbf{45.33} & 43.66 & 43.68\\
Spanish & $\checkmark$ & $\checkmark$ & $\checkmark$ & $\checkmark$ & 55.27 & 55.35 & 55.00 & \textbf{55.79} & 55.25 & 53.11 & 52.78\\
Swahili & $\checkmark$ & $\checkmark$ & $\checkmark$ & -- & \textbf{43.36} & 42.73 & 41.70 & 43.15 & 43.34 & 42.95 & 42.74\\
Thai & $\checkmark$ & $\checkmark$ & -- & -- & 37.74 & 37.94 & \textbf{38.55} & 38.40 & 31.66 & 34.57 & 31.82\\
Turkish & $\checkmark$ & $\checkmark$ & -- & -- & 38.01 & 38.98 & 39.06 & \textbf{40.57} & 40.16 & 37.02 & 37.54\\
Urdu & $\checkmark$ & $\checkmark$ & -- & -- & 37.26 & \textbf{37.77} & 36.44 & 37.29 & 36.50 & 34.91 & 36.40\\
Vietnamese & $\checkmark$ & $\checkmark$ & -- & -- & 38.90 & 40.57 & 39.89 & \textbf{41.40} & 40.74 & 37.86 & 38.46\\

\bottomrule
\end{tabular}

\caption{Aggregate downstream accuracy (\%) for languages covered by at least 2 of the four multilingual benchmarks. A checkmark identifies the benchmarks included for each language. For each tokenizer and language, accuracy is averaged equally across the checked tasks within each fine-tuning seed and then averaged over five seeds. All configurations use one pretrained encoder seed and a 200k vocabulary. The best result in each row is bold.}

\label{tab:fw104-multibenchmark-languages}
\end{table*}

\paragraph{Ranking criterion}

We evaluate 48 LCT configurations, comprising 4 structural variants, 6 values of $\tau$, and 2 vocabulary sizes, alongside BPE, Unigram, and Parity-Aware BPE. At each vocabulary size, configurations are ranked using fertility and MorphScore. Because the two metrics are not directly commensurable, we rank them separately and average the ranks. Fertility measures tokens per word, while MorphScore measures alignment with morpheme boundaries. Writing $\mathcal{S}$ for the systems evaluated at one vocabulary size, $F(s)$ for fertility, and $M(s)$ for MorphScore,

\begin{align}
r_F(s) &= 1 + \bigl|\{\,s' \in \mathcal{S} : F(s') < F(s)\,\}\bigr|, \\
r_M(s) &= 1 + \bigl|\{\,s' \in \mathcal{S} : M(s') > M(s)\,\}\bigr|,
\label{eq:metric-rank}
\end{align}

and systems are ordered by

\begin{equation}
R(s) = \tfrac{1}{2}\bigl(r_F(s) + r_M(s)\bigr),
\label{eq:mean-rank}
\end{equation}

with lower values preferred and fertility breaking ties. Fertility determines sequence length and therefore affects both computational cost and how much text fits within the context window, while MorphScore reflects whether token boundaries align with morpheme boundaries, which the encoder would otherwise need to recover. We weight the two criteria equally as a design choice rather than deriving the weighting from the data.

The ranking is stable at the top, with \texttt{LCT-MEM} at $\tau=0.2$ ranking first and $\tau=0$ second at both vocabulary sizes. \texttt{LCT-MEM} gives the strongest MorphScore, while \texttt{LCT-MDL} variants give the shortest sequences. Increasing the vocabulary from 200k to 256k lowers fertility and raises MorphScore for every displayed \lct configuration, but does not change this central trade-off. The best-ranked configurations also occur at $\tau\leq0.4$, indicating that strong or moderate language balancing is more useful than raw-frequency weighting when jointly optimizing these two criteria. Gini changes much less than fertility or MorphScore, with \texttt{LCT-MEM} at $\tau=0$ performing best at 200k and \bpe performing best at 256k among the displayed configurations. Vocabulary growth therefore improves compression and morphological alignment without automatically making cross-language tokenization cost more even.

Next, we select four \lct configurations for component-level encoder pretraining: the top-ranked \texttt{LCT-MEM} configurations at $\tau=0.2$ and $\tau=0$, and the lighter \texttt{LCT-MDL} configurations at $\tau=0$ and $\tau=0.2$. Although \texttt{LCT-MDL} ranks lower overall, it achieves the best fertility and omits branching-entropy search and morphotactic induction. Comparing these pairs tests whether the additional structural components in \texttt{LCT-MEM} improve encoder representations beyond what can be explained by shorter sequences alone, while BPE, Parity-Aware BPE, and Unigram provide external baselines.

\definecolor{LCTspread}{gray}{0.93}

\begin{table*}[h]
\centering
\small
\setlength{\tabcolsep}{4pt}

\begin{tabular}{lrrrrrrrr}
\toprule
Tokenizer & Score &
\shortstack{\texttt{LCT-MDL}\\$\tau=0$} &
\shortstack{\texttt{LCT-MDL}\\$\tau=0.2$} &
\shortstack{\texttt{LCT-MEM}\\$\tau=0$} &
\shortstack{\texttt{LCT-MEM}\\$\tau=0.2$} &
BPE &
Unigram &
\shortstack{Parity-Aware\\BPE} \\
\midrule

\shortstack{\texttt{LCT-MDL} $\tau=0$} &
$32.84 \pm 0.38$ &
\cellcolor{LCTspread} &
\cellcolor{LCTspread}$-0.80^{*}$ &
\cellcolor{LCTspread}$-0.24$ &
\cellcolor{LCTspread}$\mathbf{-1.45}^{**,\ddagger}$ &
\cellcolor{LCTspread}$+0.03$ &
\cellcolor{LCTspread}$+0.38$ &
\cellcolor{LCTspread}$+0.55$ \\

\shortstack{\texttt{LCT-MDL} $\tau=0.2$} &
$33.64 \pm 0.32$ &
\cellcolor{LCTspread}$0.54$ &
\cellcolor{LCTspread} &
\cellcolor{LCTspread}$+0.56$ &
\cellcolor{LCTspread}$-0.65^{*}$ &
\cellcolor{LCTspread}$+0.83^{*}$ &
\cellcolor{LCTspread}$+1.18^{**}$ &
\cellcolor{LCTspread}$\mathbf{+1.35}^{**}$ \\

\shortstack{\texttt{LCT-MEM} $\tau=0$} &
$33.08 \pm 0.64$ &
\cellcolor{LCTspread}$0.65$ &
\cellcolor{LCTspread}$0.78$ &
\cellcolor{LCTspread} &
\cellcolor{LCTspread}$-1.21^{*}$ &
\cellcolor{LCTspread}$+0.27$ &
\cellcolor{LCTspread}$+0.62$ &
\cellcolor{LCTspread}$+0.79$ \\

\shortstack{\texttt{LCT-MEM} $\tau=0.2$} &
$34.29 \pm 0.33$ &
\cellcolor{LCTspread}$0.23$ &
\cellcolor{LCTspread}$0.35$ &
\cellcolor{LCTspread}$0.70$ &
\cellcolor{LCTspread} &
\cellcolor{LCTspread}$\mathbf{+1.48}^{**,\dagger}$ &
\cellcolor{LCTspread}$\mathbf{+1.83}^{**,\dagger}$ &
\cellcolor{LCTspread}$\mathbf{+2.00}^{**}$ \\

BPE &
$32.81 \pm 0.26$ &
\cellcolor{LCTspread}$0.39$ &
\cellcolor{LCTspread}$0.51$ &
\cellcolor{LCTspread}$0.40$ &
\cellcolor{LCTspread}$0.43$ &
\cellcolor{LCTspread} &
\cellcolor{LCTspread}$+0.35$ &
\cellcolor{LCTspread}$+0.52$ \\

Unigram &
$32.46 \pm 0.22$ &
\cellcolor{LCTspread}$0.38$ &
\cellcolor{LCTspread}$0.50$ &
\cellcolor{LCTspread}$0.77$ &
\cellcolor{LCTspread}$0.42$ &
\cellcolor{LCTspread}$0.38$ &
\cellcolor{LCTspread} &
\cellcolor{LCTspread}$+0.17$ \\

\shortstack{Parity-Aware\\BPE} &
$32.29 \pm 0.61$ &
\cellcolor{LCTspread}$0.94$ &
\cellcolor{LCTspread}$0.62$ &
\cellcolor{LCTspread}$1.03$ &
\cellcolor{LCTspread}$0.85$ &
\cellcolor{LCTspread}$0.75$ &
\cellcolor{LCTspread}$0.68$ &
\cellcolor{LCTspread} \\

\bottomrule
\end{tabular}

\caption{Pairwise comparison of the aggregated downstream score (accuracy \%). Score is each configuration's mean and standard deviation over five fine-tuning seeds. Rows and columns contain the same seven configurations in the same order. All configurations use a 200k vocabulary and one pretrained encoder each. Values above the diagonal report the paired difference between the row and column configurations, with positive values favouring the row. 
$^{*}$ and $^{**}$ indicate raw paired-test $p<0.05$ and $p<0.01$, while $^{\dagger}$ and $^{\ddagger}$ indicate Holm-adjusted $p<0.05$ and $p<0.01$, respectively. Values that remain significant after correction are bold.}

\label{tab:fw104-pairwise}
\end{table*}

\subsection{Extrinsic Evaluation}
\label{sec:extrinsic-results}

All seven encoders use the same architecture and pretraining seed, and each is evaluated with the same five fine-tuning seeds; only the tokenizer differs.

Table~\ref{tab:fw104-pairwise} separates the effects of language balancing and structural objective. Changing only $\tau$ improves \texttt{LCT-MDL} by $+0.80$ points and \texttt{LCT-MEM} by $+1.21$ points, showing that balancing strength is a meaningful model-selection parameter rather than a fixed preprocessing choice. At $\tau=0.2$, \texttt{LCT-MEM} improves by $+0.65$ points over \texttt{LCT-MDL} in the raw paired test. \texttt{LCT-MDL} nevertheless remains useful because it yields the shortest sequences, omits branching-entropy search and morphotactic induction, and still outperforms BPE. \texttt{LCT-MEM}, in contrast, provides the strongest downstream performance at the cost of a more complex tokenizer-construction pipeline.

The gains are not uniform across resource levels. On Belebele, \texttt{LCT-MEM} at $\tau=0.2$ is effectively tied with BPE in the smallest pretraining-corpus quartile ($-0.07$ points), but improves by $+2.45$, $+2.58$, and $+1.95$ points in the remaining quartiles. Better vocabulary allocation therefore does not by itself guarantee better outcomes for languages with the least pretraining data, indicating that data availability remains a separate bottleneck.

\section{Discussion}
\label{sec:discussion}

\subsection{A controllable tokenizer design space}

Standard BPE and Unigram couple corpus frequency, unit discovery, and surface vocabulary construction inside one objective. \lct exposes two of these choices as independent controls. The temperature $\tau$ determines how strongly corpus size influences multilingual evidence, while the discovery objective determines how much structural information is preserved before surface packing. This makes vocabulary allocation a tunable design decision rather than an incidental consequence of the corpus distribution.

The downstream ordering shows why this control matters. Complete equalization at $\tau=0$ is not consistently better than BPE, whereas partial balancing at $\tau=0.2$ improves both \texttt{LCT-MDL} and \texttt{LCT-MEM}. The useful operating point therefore lies between unrestricted raw frequency and equal language totals. With only two pretrained temperatures, we identify a direction rather than an optimum, but the results show that allocation strength can be selected empirically for the target setting.

\subsection{Different objectives support different deployment priorities}

The component choice provides a second operating axis. \texttt{LCT-MEM} gives the strongest morphological alignment and downstream performance, while \texttt{LCT-MDL} produces shorter sequences and omits branching-entropy search and morphotactic induction. Figure~\ref{fig:resource-intrinsic} shows that \texttt{LCT-MEM} maintains higher MorphScore than the baselines across low-, mid-, and high-resource groups, while vocabulary utilization is also consistently higher or comparable. Fertility is more resource dependent, with clearer LCT gains for low-resource languages and smaller differences relative to BPE in the mid- and high-resource groups. The intrinsic advantage of LCT is therefore not explained by compression alone.

This should not be confused with a demonstrated tokenizer-runtime speedup. Shorter sequences reduce encoder computation, and \texttt{LCT-MDL} removes training-time components, but the current \lct Hugging Face adapter is implemented as a Python slow tokenizer. Compiling the lossless packing algorithm into an optimized runtime remains necessary before making serving-latency claims.

\subsection{Allocation parity is not outcome parity}

The downstream gains are also resource dependent. Figure~\ref{fig:resource-downstream} shows relatively small differences among tokenizers for low-resource languages, with clearer separation in the mid- and high-resource groups, where \texttt{LCT-MEM} at $\tau=0.2$ performs strongest. This suggests that better structural units and vocabulary allocation are most useful when the encoder has sufficient pretraining data to learn from them.

The result also exposes an important limitation. \lct can redistribute vocabulary capacity and reduce cross-language fragmentation, but better allocation does not by itself guarantee stronger outcomes for languages with the least pretraining data. Languages with very little training text may receive better pieces without providing enough evidence for the encoder to exploit them. Future work should therefore optimize low-resource outcomes directly rather than treating lower allocation or Gini as sufficient, including resource-aware temperature schedules, language- or script-conditioned allocation floors, downstream-sensitive structural objectives, and evaluation with larger low-resource task suites and independent pretraining seeds.

\section{Related Work}
\label{sec:related_work}

Multilingual tokenization must distribute finite vocabulary capacity across languages with highly unequal data availability. Prior work has studied language specific vocabulary allocation, cross language vocabulary overlap, and disparities in tokenization efficiency \citep{rust-etal-2021-good,zheng-etal-2021-allocating,limisiewicz-etal-2023-tokenization,lundin-etal-2026-token}. Parity Aware BPE directly modifies BPE merge selection to reduce cross language compression disparities \citep{foroutan-etal-2026-parity}. LCT similarly reduces the influence of corpus size on multilingual vocabulary learning, but separates language balancing and latent structural discovery from subsequent surface vocabulary construction.

A complementary line of work incorporates morphological structure into tokenization. Unsupervised morphological segmentation has long used minimum description length to identify reusable subword structure \citep{10.1145/1187415.1187418}, while recent approaches incorporate morphological information through boundary constraints, segmented training data, or transformations that expose internal structure \citep{asgari-etal-2026-morphbpe,garcia-etal-2025-exploring,vemula-etal-2025-rethinking,gazit-etal-2025-splintering}. MorphBPE is particularly close to LCT in using morphology during vocabulary learning, but its morphological boundaries constrain the final BPE segmentation. LCT instead uses structural analysis as an intermediate latent representation and allows latent pieces to be recombined when constructing the final surface vocabulary.

Recent work also shows that compression, morphological alignment, and downstream performance need not favor the same tokenizer \citep{vemula-etal-2025-rethinking,asgari-etal-2026-morphbpe}. LCT builds on this distinction by separating structural discovery from surface vocabulary construction and evaluating both tokenizer quality and learned multilingual representations.

\section{Conclusion}
\label{sec:conclusion}

We introduce the Latent Core Tokenizer (LCT), which separates latent structural discovery from surface vocabulary construction. Across 104 languages, LCT improves intrinsic tokenizer quality over BPE, Unigram, and Parity-Aware BPE, while controlled encoder pretraining shows that \texttt{LCT-MEM} at $\tau=0.2$ improves aggregate downstream performance by 1.48, 1.83, and 2.00 points over the three baselines, respectively. Our ablations further show that the configuration producing the shortest sequences is not the one producing the strongest downstream representations, indicating that compression alone is not sufficient for tokenizer selection. The gains also vary by resource level, with smaller downstream improvements for languages with the least pretraining data, showing that better tokenization can improve multilingual representation and vocabulary allocation without removing the separate constraint imposed by limited data.

\section*{Limitations}
\label{sec:limitations}

Our results provide evidence that separating structural discovery from surface vocabulary construction can improve multilingual tokenization and downstream representations, but several aspects of the current evaluation limit the scope of these conclusions.

\emph{Single pretraining seed}: all seven encoders were pretrained once, so the reported variance reflects fine-tuning variation only. We therefore cannot exclude the possibility that part of the $1.48$-point improvement over BPE reflects pretraining variance. Replicating encoder pretraining across multiple seeds would provide a stronger estimate of the tokenizer effect.

\emph{Scale}: our experiments use DeBERTa-v3 XSmall encoders. The observed effects therefore establish the value of LCT at this model scale, but may not transfer unchanged to larger models, which may be better able to recover from suboptimal segmentation.

\emph{Task coverage}: 62 of the 79 evaluated languages are represented only in Belebele, making the aggregate result heavily influenced by this benchmark. XNLI, XStoryCloze, and PAWS-X cover substantially fewer languages, and XStoryCloze has only 360 training examples. Broader multilingual evaluation is therefore needed to determine how consistently the observed gains extend across task types.

\emph{Search}: although 48 LCT configurations are evaluated intrinsically, only two values of $\tau$ and two structural objectives are taken forward to encoder pretraining. The downstream experiments therefore identify useful directions rather than an optimal configuration, and conclusions about the relationship between intrinsic and downstream behavior are based on a much smaller subset of the full design space.

\emph{Single downstream architecture}: all downstream experiments use the same DeBERTa-v3 XSmall architecture. This controls architecture as a source of variation, but does not establish that the observed tokenizer effects generalize to other encoder architectures or to decoder-only language models.

\section*{Ethical Considerations}
\label{sec:ethics}

This work does not involve human participants or the collection of new personal data. The experiments use existing multilingual corpora and public evaluation benchmarks. 


\bibliography{custom}

@inproceedings{sennrich-etal-2016-neural,
    title = "Neural Machine Translation of Rare Words with Subword Units",
    author = "Sennrich, Rico  and
      Haddow, Barry  and
      Birch, Alexandra",
    editor = "Erk, Katrin  and
      Smith, Noah A.",
    booktitle = "Proceedings of the 54th Annual Meeting of the Association for Computational Linguistics (Volume 1: Long Papers)",
    month = aug,
    year = "2016",
    address = "Berlin, Germany",
    publisher = "Association for Computational Linguistics",
    url = "https://aclanthology.org/P16-1162/",
    doi = "10.18653/v1/P16-1162",
    pages = "1715--1725"
}

@inproceedings{kudo-2018-subword,
    title = "Subword Regularization: Improving Neural Network Translation Models with Multiple Subword Candidates",
    author = "Kudo, Taku",
    editor = "Gurevych, Iryna  and
      Miyao, Yusuke",
    booktitle = "Proceedings of the 56th Annual Meeting of the Association for Computational Linguistics (Volume 1: Long Papers)",
    month = jul,
    year = "2018",
    address = "Melbourne, Australia",
    publisher = "Association for Computational Linguistics",
    url = "https://aclanthology.org/P18-1007/",
    doi = "10.18653/v1/P18-1007",
    pages = "66--75"
}

@inproceedings{rust-etal-2021-good,
    title = "How Good is Your Tokenizer? On the Monolingual Performance of Multilingual Language Models",
    author = "Rust, Phillip  and
      Pfeiffer, Jonas  and
      Vuli{\'c}, Ivan  and
      Ruder, Sebastian  and
      Gurevych, Iryna",
    editor = "Zong, Chengqing  and
      Xia, Fei  and
      Li, Wenjie  and
      Navigli, Roberto",
    booktitle = "Proceedings of the 59th Annual Meeting of the Association for Computational Linguistics and the 11th International Joint Conference on Natural Language Processing (Volume 1: Long Papers)",
    month = aug,
    year = "2021",
    address = "Online",
    publisher = "Association for Computational Linguistics",
    url = "https://aclanthology.org/2021.acl-long.243/",
    doi = "10.18653/v1/2021.acl-long.243",
    pages = "3118--3135"
}

@inproceedings{lundin-etal-2026-token,
    title = "The Token Tax: Systematic Bias in Multilingual Tokenization",
    author = "Lundin, Jessica M.  and
      Zhang, Ada  and
      Karim, Nihal  and
      Louzan, Hamza  and
      Wei, Guohao  and
      Adelani, David Ifeoluwa  and
      Carroll, Cody",
    editor = "Chimoto, Everlyn Asiko  and
      Lignos, Constantine  and
      Muhammad, Shamsuddeen  and
      Abdulmumin, Idris  and
      Siro, Clemencia  and
      Adelani, David Ifeoluwa",
    booktitle = "Proceedings of the 7th Workshop on {A}frican Natural Language Processing ({A}frica{NLP} 2026)",
    month = mar,
    year = "2026",
    address = "Rabat, Morocco",
    publisher = "Association for Computational Linguistics",
    url = "https://aclanthology.org/2026.africanlp-main.10/",
    doi = "10.18653/v1/2026.africanlp-main.10",
    pages = "103--112",
    ISBN = "979-8-89176-364-7"
}

@inproceedings{foroutan-etal-2026-parity,
    title = "Parity-Aware Byte-Pair Encoding: Improving Cross-lingual Fairness in Tokenization",
    author = "Foroutan, Negar  and
      Meister, Clara  and
      Paul, Debjit  and
      Niklaus, Joel  and
      Ahmadi, Sina  and
      Bosselut, Antoine  and
      Sennrich, Rico",
    editor = "Liakata, Maria  and
      Moreira, Viviane P.  and
      Zhang, Jiajun  and
      Jurgens, David",
    booktitle = "Proceedings of the 64th Annual Meeting of the {A}ssociation for {C}omputational {L}inguistics (Volume 1: Long Papers)",
    month = jul,
    year = "2026",
    address = "San Diego, California, United States",
    publisher = "Association for Computational Linguistics",
    url = "https://aclanthology.org/2026.acl-long.342/",
    doi = "10.18653/v1/2026.acl-long.342",
    pages = "7514--7538",
    ISBN = "979-8-89176-390-6"
}

@article{10.1145/1187415.1187418,
author = {Creutz, Mathias and Lagus, Krista},
title = {Unsupervised models for morpheme segmentation and morphology learning},
year = {2007},
issue_date = {January 2007},
publisher = {Association for Computing Machinery},
address = {New York, NY, USA},
volume = {4},
number = {1},
issn = {1550-4875},
url = {https://doi.org/10.1145/1187415.1187418},
doi = {10.1145/1187415.1187418},
journal = {ACM Trans. Speech Lang. Process.},
month = feb,
articleno = {3},
numpages = {34}
}

@inproceedings{asgari-etal-2026-morphbpe,
    title = "{M}orph{BPE}: Morphology-Aware Tokenization for Efficient {LLM} Training",
    author = "Asgari, Ehsaneddin  and
      El Kheir, Yassine  and
      SadraeiJavaheri, MohammadAli  and
      Nazari, Ali",
    editor = "Liakata, Maria  and
      Moreira, Viviane P.  and
      Zhang, Jiajun  and
      Jurgens, David",
    booktitle = "Findings of the {A}ssociation for {C}omputational {L}inguistics: {ACL} 2026",
    month = jul,
    year = "2026",
    address = "San Diego, California, United States",
    publisher = "Association for Computational Linguistics",
    url = "https://aclanthology.org/2026.findings-acl.2068/",
    doi = "10.18653/v1/2026.findings-acl.2068",
    pages = "41610--41621",
    ISBN = "979-8-89176-395-1"
}

@inproceedings{zheng-etal-2021-allocating,
    title = "Allocating Large Vocabulary Capacity for Cross-Lingual Language Model Pre-Training",
    author = "Zheng, Bo  and
      Dong, Li  and
      Huang, Shaohan  and
      Singhal, Saksham  and
      Che, Wanxiang  and
      Liu, Ting  and
      Song, Xia  and
      Wei, Furu",
    editor = "Moens, Marie-Francine  and
      Huang, Xuanjing  and
      Specia, Lucia  and
      Yih, Scott Wen-tau",
    booktitle = "Proceedings of the 2021 Conference on Empirical Methods in Natural Language Processing",
    month = nov,
    year = "2021",
    address = "Online and Punta Cana, Dominican Republic",
    publisher = "Association for Computational Linguistics",
    url = "https://aclanthology.org/2021.emnlp-main.257/",
    doi = "10.18653/v1/2021.emnlp-main.257",
    pages = "3203--3215"
}

@inproceedings{vemula-etal-2025-rethinking,
    title = "Rethinking Tokenization for Rich Morphology: The Dominance of Unigram over {BPE} and Morphological Alignment",
    author = "Vemula, Saketh Reddy  and
      Dandapat, Sandipan  and
      Sharma, Dipti  and
      Krishnamurthy, Parameswari",
    editor = "T.Y.S.S, Santosh  and
      Shimizu, Shuichiro  and
      Gong, Yifan",
    booktitle = "The 14th International Joint Conference on Natural Language Processing and The 4th Conference of the Asia-Pacific Chapter of the Association for Computational Linguistics",
    month = dec,
    year = "2025",
    address = "Mumbai, India",
    publisher = "Association for Computational Linguistics",
    url = "https://aclanthology.org/2025.ijcnlp-srw.20/",
    doi = "10.18653/v1/2025.ijcnlp-srw.20",
    pages = "232--252",
    ISBN = "979-8-89176-304-3"
}

@inproceedings{limisiewicz-etal-2023-tokenization,
    title = "Tokenization Impacts Multilingual Language Modeling: Assessing Vocabulary Allocation and Overlap Across Languages",
    author = "Limisiewicz, Tomasz  and
      Balhar, Ji{\v{r}}{\'i}  and
      Mare{\v{c}}ek, David",
    editor = "Rogers, Anna  and
      Boyd-Graber, Jordan  and
      Okazaki, Naoaki",
    booktitle = "Findings of the Association for Computational Linguistics: ACL 2023",
    month = jul,
    year = "2023",
    address = "Toronto, Canada",
    publisher = "Association for Computational Linguistics",
    url = "https://aclanthology.org/2023.findings-acl.350/",
    doi = "10.18653/v1/2023.findings-acl.350",
    pages = "5661--5681"
}

@inproceedings{garcia-etal-2025-exploring,
    title = "Exploring morphology-aware tokenization: A case study on {S}panish language modeling",
    author = "Garc{\'i}a, Alba T{\'a}boas  and
      Przyby{\l}a, Piotr  and
      Wanner, Leo",
    editor = "Christodoulopoulos, Christos  and
      Chakraborty, Tanmoy  and
      Rose, Carolyn  and
      Peng, Violet",
    booktitle = "Proceedings of the 2025 Conference on Empirical Methods in Natural Language Processing",
    month = nov,
    year = "2025",
    address = "Suzhou, China",
    publisher = "Association for Computational Linguistics",
    url = "https://aclanthology.org/2025.emnlp-main.1552/",
    doi = "10.18653/v1/2025.emnlp-main.1552",
    pages = "30505--30518",
    ISBN = "979-8-89176-332-6"
}

@inproceedings{gazit-etal-2025-splintering,
    title = "Splintering Nonconcatenative Languages for Better Tokenization",
    author = "Gazit, Bar  and
      Shmidman, Shaltiel  and
      Shmidman, Avi  and
      Pinter, Yuval",
    editor = "Che, Wanxiang  and
      Nabende, Joyce  and
      Shutova, Ekaterina  and
      Pilehvar, Mohammad Taher",
    booktitle = "Findings of the Association for Computational Linguistics: ACL 2025",
    month = jul,
    year = "2025",
    address = "Vienna, Austria",
    publisher = "Association for Computational Linguistics",
    url = "https://aclanthology.org/2025.findings-acl.1151/",
    doi = "10.18653/v1/2025.findings-acl.1151",
    pages = "22405--22417",
    ISBN = "979-8-89176-256-5"
}

@inproceedings{bandarkar-etal-2024-belebele,
    title = "The Belebele Benchmark: a Parallel Reading Comprehension Dataset in 122 Language Variants",
    author = "Bandarkar, Lucas  and
      Liang, Davis  and
      Muller, Benjamin  and
      Artetxe, Mikel  and
      Shukla, Satya Narayan  and
      Husa, Donald  and
      Goyal, Naman  and
      Krishnan, Abhinandan  and
      Zettlemoyer, Luke  and
      Khabsa, Madian",
    booktitle = "Proceedings of the 62nd Annual Meeting of the Association for Computational Linguistics (Volume 1: Long Papers)",
    month = aug,
    year = "2024",
    address = "Bangkok, Thailand and virtual meeting",
    publisher = "Association for Computational Linguistics",
    url = "https://aclanthology.org/2024.acl-long.44",
    pages = "749--775",
}

@inproceedings{arnett2025alignment,
  author = {Arnett, Catherine and Hudspeth, Marisa and O'Connor, Brendan},
  title = {{Evaluating Morphological Alignment of Tokenizers in 70 Languages}},
  year = {2025},
  booktitle={Proceedings of the ICML 2025 Tokenization Workshop (TokShop)},
  url = {https://arxiv.org/abs/2507.06378}
}

@inproceedings{conneau-etal-2018-xnli,
    title = "{XNLI}: Evaluating Cross-lingual Sentence Representations",
    author = "Conneau, Alexis  and
      Rinott, Ruty  and
      Lample, Guillaume  and
      Williams, Adina  and
      Bowman, Samuel R.  and
      Schwenk, Holger  and
      Stoyanov, Veselin",
    editor = "Riloff, Ellen  and
      Chiang, David  and
      Hockenmaier, Julia  and
      Tsujii, Jun{'}ichi",
    booktitle = "Proceedings of the 2018 Conference on Empirical Methods in Natural Language Processing",
    month = oct # "-" # nov,
    year = "2018",
    address = "Brussels, Belgium",
    publisher = "Association for Computational Linguistics",
    url = "https://aclanthology.org/D18-1269/",
    doi = "10.18653/v1/D18-1269",
    pages = "2475--2485"
}

@inproceedings{mihaylov-frank-2017-story,
    title = "Story Cloze Ending Selection Baselines and Data Examination",
    author = "Mihaylov, Todor  and
      Frank, Anette",
    editor = "Roth, Michael  and
      Mostafazadeh, Nasrin  and
      Chambers, Nathanael  and
      Louis, Annie",
    booktitle = "Proceedings of the 2nd Workshop on Linking Models of Lexical, Sentential and Discourse-level Semantics",
    month = apr,
    year = "2017",
    address = "Valencia, Spain",
    publisher = "Association for Computational Linguistics",
    url = "https://aclanthology.org/W17-0913/",
    doi = "10.18653/v1/W17-0913",
    pages = "87--92"
}

@inproceedings{yang-etal-2019-paws,
    title = "{PAWS}-{X}: A Cross-lingual Adversarial Dataset for Paraphrase Identification",
    author = "Yang, Yinfei  and
      Zhang, Yuan  and
      Tar, Chris  and
      Baldridge, Jason",
    editor = "Inui, Kentaro  and
      Jiang, Jing  and
      Ng, Vincent  and
      Wan, Xiaojun",
    booktitle = "Proceedings of the 2019 Conference on Empirical Methods in Natural Language Processing and the 9th International Joint Conference on Natural Language Processing (EMNLP-IJCNLP)",
    month = nov,
    year = "2019",
    address = "Hong Kong, China",
    publisher = "Association for Computational Linguistics",
    url = "https://aclanthology.org/D19-1382/",
    doi = "10.18653/v1/D19-1382",
    pages = "3687--3692"
}

\clearpage

\onecolumn
\appendix
\section{Appendix}

\subsection{LCT algorithm}

Algorithm~\ref{alg:LCT-training} summarizes how these operations are executed
to produce the latent analyzer, fixed surface vocabulary, and ranked merges.
The language-balanced table drives both latent discovery and ranked packing.
$C$ is a weighted multiset of latent piece sequences, while $H[p]$ records
the weighted support of piece $p$ across word types.  Pair gain is the current
weighted adjacent-pair count: replacing one occurrence saves one emitted
token.  Recomputing affected counts after every non-overlapping merge makes
the merge order deterministic and allows complete words to be reconstructed
from latent pieces.  If support threshold 10 cannot fill the exact budget,
the implementation retries the packing stage with threshold 1.

\begin{algorithm}[h]
\caption{\lct multilingual vocabulary training}
\label{alg:LCT-training}
\scriptsize
\begin{algorithmic}[1]
\Require Corpora $\{\mathcal{D}_{\ell}\}$; budget $K$; byte budget $B$;
base fraction $\beta$; support $m$
\Ensure Analyzer $A$, vocabulary $V$, ranked merges $R$

\ForAll{languages $\ell$}
  \State $F_{\ell}\gets\Call{CountWords}{\mathcal{D}_{\ell}}$;
  $N_{\ell}\gets\sum_wF_{\ell}(w)$
\EndFor
\State $N_{\max}\gets\max_{\ell}N_{\ell}$
\State $\widetilde F(w)\gets
\sum_{\ell:F_{\ell}(w)>0}
\max(1,\Call{Round}{F_{\ell}(w)N_{\max}/N_{\ell}})$

\State $q^\star\gets\Call{SelectEntropyThreshold}{\widetilde F}$
\State $S\gets\Call{TrainMDLSegmenter}{\widetilde F,q^\star}$
\State $M\gets\Call{FitMorphotactics}{S,\widetilde F}$
\State $A\gets\Call{AddCoreAndAffixMerges}{S,M,\widetilde F}$

\State $C,H\gets\Call{EmptyWeightedCounts}{}$
\ForAll{$(w,f)\in\widetilde F$}
  \State $z\gets\Call{LatentTokenize}{A,w}$
  \State $C[z]\mathrel{+}=f$;
  $\Call{AddDistinctPieceSupport}{H,z,f}$
\EndFor

\State $K_{\mathrm{learned}}\gets K-B$
\State $V\gets\Call{CharacterAlphabet}{\widetilde F}$
\State $K_{\mathrm{base}}\gets
\max(|V|,\lfloor\beta K_{\mathrm{learned}}\rfloor)$
\State Add highest $H[p]$ pieces with $H[p]\ge m$ until $|V|=K_{\mathrm{base}}$

\State $m_{\mathrm{eff}}\gets m$
\While{$|V|<K_{\mathrm{learned}}$}
  \State $(a,b),g\gets\Call{HighestWeightedAdjacentGain}{C}$
  \If{$g<m_{\mathrm{eff}}$}
    \If{$m_{\mathrm{eff}}>1$}
      \State $m_{\mathrm{eff}}\gets1$
    \Else
      \State \textbf{fail}
    \EndIf
  \Else
    \State $\Call{MergeNonOverlapping}{C,(a,b)\mapsto ab}$
    \State $V\gets V\cup\{ab\}$;
    $\Call{Append}{R,(a,b)}$
  \EndIf
\EndWhile
\State $V\gets V\cup\{\textsc{Byte}(0),\ldots,\textsc{Byte}(255)\}$
\State \Return $(A,V,R)$
\end{algorithmic}
\end{algorithm}

\subsubsection{Lossless runtime}
\label{sec:runtime}

\paragraph{Protected text segmentation.}
Whitespace and punctuation are preserved rather than discarded.  A synthetic
boundary marker distinguishes word-initial pieces while remaining reversible.
Punctuation-aware vocabulary allocation reduces repeated byte fallback for
common layout symbols.

\paragraph{Known-piece backoff.}
The latent analyzer can produce a surface piece that was pruned from the fixed
vocabulary.  Instead of immediately replacing the entire piece with bytes,
the runtime uses dynamic programming to find the fewest known vocabulary
pieces whose concatenation exactly reconstructs it.  Only irreducible spans
fall back to one of 256 UTF-8 byte IDs.

\paragraph{Minimum-segment packing.}
The corrected runtime optimizes each complete protected segment rather than
accepting the latent analyzer's pieces independently.  For character position
$i$, it evaluates every known vocabulary piece beginning at $i$ and the UTF-8
byte encoding of the current character.  Paths are ordered by:
\begin{enumerate}[leftmargin=*]
  \item total number of emitted IDs;
  \item number of byte IDs;
  \item preference for learned pieces over byte fallback; and
  \item preference for longer learned pieces.
\end{enumerate}
Backtracking returns the globally minimum-token exact encoding for that
segment.  Decoding concatenates learned surfaces and buffered UTF-8 bytes,
then restores protected whitespace.  Round-trip equality is enforced during
corpus preparation.

\begin{algorithm}[t]
\caption{Lossless minimum-segment encoding}
\label{alg:minimum-packing}
\begin{algorithmic}[1]
\Require Protected segment $s=s_0\ldots s_{n-1}$; known learned pieces $V$
\Ensure Exact token sequence with minimum lexicographic path cost
\State $T[n]\gets0$; $B[n]\gets0$
  \Comment{total-token and byte-token suffix costs}
\For{$i=n-1$ downto $0$}
  \State $u\gets\Call{UTF8Bytes}{s_i}$
  \State $k^\star\gets
    (|u|+T[i+1],\,|u|+B[i+1],\,1,\,0)$
  \State $c^\star\gets(\Call{ByteTokens}{u},i+1)$
  \ForAll{lengths $d$ of known pieces, from longest to shortest}
    \State $j\gets i+d$
    \If{$j\le n$ and $s_{i:j}\in V$}
      \State $k\gets(1+T[j],\,B[j],\,0,\,-d)$
      \If{$k<_{\mathrm{lex}}k^\star$}
        \State $k^\star\gets k$; $c^\star\gets((s_{i:j}),j)$
      \EndIf
    \EndIf
  \EndFor
  \State $T[i]\gets k^\star_1$; $B[i]\gets k^\star_2$;
    $\mathrm{choice}[i]\gets c^\star$
\EndFor
\State $i\gets0$; $y\gets[\,]$
\While{$i<n$}
  \State $(q,j)\gets\mathrm{choice}[i]$; append $q$ to $y$; $i\gets j$
\EndWhile
\State \Return $y$
\end{algorithmic}
\end{algorithm}

Algorithm~\ref{alg:minimum-packing} uses a four-part lexicographic cost:
total IDs, byte IDs, a learned-piece preference, and negative piece length.
Thus it first minimizes sequence length, then byte fallback, then resolves
ties in favor of learned and longer pieces.  Every character always has a
UTF-8 fallback edge, so a complete path exists for every Unicode string.
With a hash set of known pieces and a precomputed set of distinct piece
lengths, a segment of length $n$ takes
$O(n|\mathcal{L}|)$ membership checks and $O(n)$ dynamic-programming memory.

\subsubsection{Worked example with a dummy corpus}
\label{sec:dummy-example}

Consider two small corpora with the following word counts:
\begin{center}
\begin{tabular}{llrr}
\toprule
Language & Word & Raw count & Balanced count\\
\midrule
$\ell_A$ & \texttt{walk} & 6 & 6\\
 & \texttt{walking} & 4 & 4\\
 & \texttt{walked} & 2 & 2\\
$\ell_B$ & \texttt{play} & 2 & 6\\
 & \texttt{playing} & 1 & 3\\
 & \texttt{played} & 1 & 3\\
\bottomrule
\end{tabular}
\end{center}
Here $N_A=12$, $N_B=4$, and $N_{\max}=12$, so the second language receives
multiplier three.  Before balancing, the \texttt{walk} family supplies 75\%
of all observations; afterward, the \texttt{walk} and \texttt{play} families
each supply 12 effective observations.  The shared suffixes \texttt{ing} and
\texttt{ed} can therefore receive evidence from both families without the
smaller corpus being overwhelmed.

For illustration, compare storing all six words with analyzing
\[
\{\texttt{walk},\texttt{play},\texttt{ing},\texttt{ed}\}.
\]
The character alphabet has 11 symbols, giving
$C_\Sigma=1.5\log_2(12)=5.377$ bits per stored character-plus-terminator unit.
The whole-word lexicon has 40 such units and corpus counts
$(6,4,2,6,3,3)$, yielding approximately
\[
\mathcal{L}_{\mathrm{whole}}
=40(5.377)+59.510=274.608\ \text{bits}.
\]
The segmented lexicon has only 17 units.  Its piece counts are
$(12,12,7,5)$ for
(\texttt{walk}, \texttt{play}, \texttt{ing}, \texttt{ed}), giving
\[
\mathcal{L}_{\mathrm{segmented}}
=17(5.377)+68.817=160.234\ \text{bits}.
\]
Segmentation increases the corpus code by introducing more symbols, but the
reusable four-piece lexicon more than compensates.  The actual trainer
evaluates incremental costs while repeatedly updating the counts; these
numbers simplify that process only to expose the trade-off.

Branching entropy supports the same stem boundaries.  After context
\texttt{walk}, the observed next-symbol distribution is
\[
P(\mathrm{END},\texttt{i},\texttt{e})
=(6,4,2)/12,
\]
whose forward entropy is 1.459 bits, or 0.407 after normalization by
$\log_2(12)$.  Inside the stem, after \texttt{wa}, the next character is
always \texttt{l}, so the corresponding entropy is zero.  The backward model
provides complementary evidence from the suffix side, and the selected
threshold determines whether the high-scoring \texttt{walk|ing} and
\texttt{walk|ed} positions are forced.

Morphotactics then labels \texttt{walk} and \texttt{play} as
\textsc{STM}, and \texttt{ing} and \texttt{ed} as \textsc{SUF}, producing
the legal patterns \textsc{STM--SUF}.  If the base surface vocabulary keeps
these four pieces, the initial weighted adjacent scores are
\[
g(\texttt{walk},\texttt{ing})=4,\quad
g(\texttt{play},\texttt{ing})=3,\quad
g(\texttt{play},\texttt{ed})=3,\quad
g(\texttt{walk},\texttt{ed})=2.
\]
With one additional surface slot, the first ranked merge creates
\texttt{walking}; further slots can create common whole forms such as
\texttt{played}.  The latent boundaries remain available during discovery,
while the final encoder receives the shorter deterministic surface
representation.

\clearpage
\section{Data}\label{app:data}

\begin{figure}[h]
    \centering
    \includegraphics[width=0.9\linewidth]{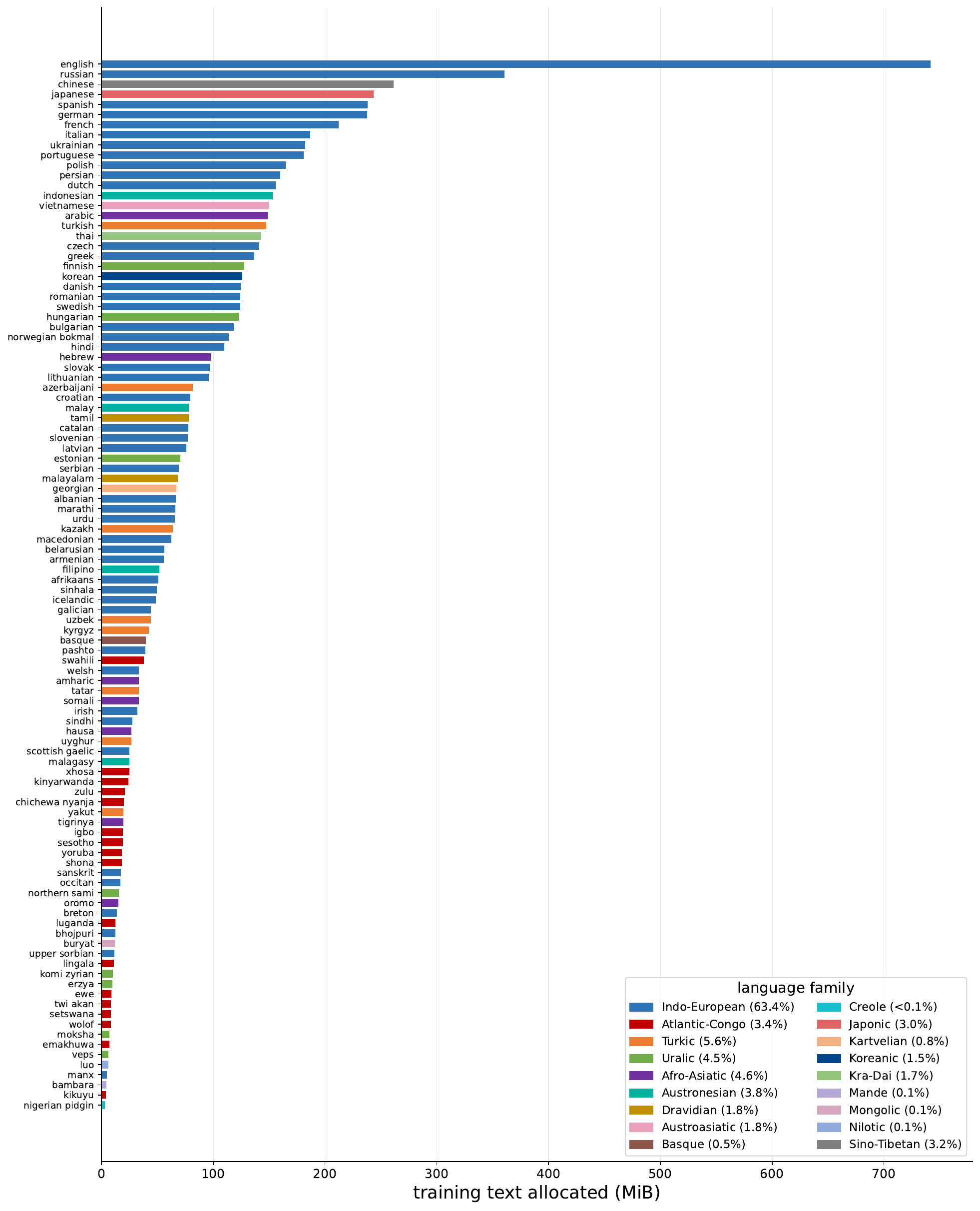}
    \caption{Per-language pretraining statistics}
    \label{fig:pretraining-language-distribution}
\end{figure}

\clearpage

\section{Detailed experimental settings}
\label{app:settings}

\subsection{Encoder}
Each tokenizer is used to train a DeBERTa-v3-style XSmall encoder with 12
discriminator layers, hidden size 384, six attention heads, intermediate size
1,536, and a maximum sequence length of 512. A six-layer generator corrupts
15\% of eligible tokens, and training uses replaced token detection:
\begin{equation}
\mathcal{L}_{\mathrm{RTD}}
=
\mathcal{L}_{\mathrm{generator}}
+
\lambda\mathcal{L}_{\mathrm{discriminator}}.
\end{equation}

\begin{table}[h]
\centering
\small
\begin{tabular}{ll}
\toprule
Component & Setting\\
\midrule
\multicolumn{2}{l}{\emph{Corpus}}\\
Pretraining corpus & FineWeb-2, 104 languages; 41,347,986 lines; 8.55\,GB\\
\quad shared by all arms & identical SHA-256; 3,946,024 packed sequences\\
Tokenizer training corpus & the same 41,347,986 lines for every arm\\

\midrule
\multicolumn{2}{l}{\emph{Encoder}}\\
Objective & DeBERTa-v3 replaced-token detection\\
Encoder & 12 layers; hidden 384; 6 heads; FFN 1,536\\
Generator & 6 layers; 15\% corruption\\
Sequence length & 512\\
$\lambda$ & 10 \\
\midrule
\multicolumn{2}{l}{\emph{Optimization}}\\
Schedule & 15k updates; LR $3\!\times\!10^{-4}$; 1k warmup; BF16\\
Batch & 8 per device $\times$ 4 accumulation $\times$ 8 GPUs = 256\\
Tokens seen & $1.97\times10^{9}$ ($\approx$ 1 epoch)\\
Compute & 8 $\times$ A100 40\,GB (NDv4) per arm\\
\midrule
\multicolumn{2}{l}{\emph{Protocol}}\\
Pretraining seeds & 1 (42)\\

Fine-tuning seeds & 5 (11, 29, 47, 53, 71)\\
Learning-rate selection & swept once on seed 11, replicated on the rest\\
\bottomrule
\end{tabular}
\caption{Encoder settings}
\label{tab:encoder-settings}
\end{table}

\subsection{Tokenizers}

\begin{table}[h]
\centering
\small
\begin{tabular}{lll}
\toprule
System & Selection criterion & Distinguishing setting\\
\midrule
LCT-MDL $\tau{=}0$ & description length & uniform language weights\\
LCT-MDL $\tau{=}0.2$ & description length & tempered weights\\
LCT-\textsc{MEM} $\tau{=}0$ & {+} entropy, morphotactics & uniform language weights\\
LCT-\textsc{MEM} $\tau{=}0.2$ & {+} entropy, morphotactics & tempered weights\\
BPE & global pair frequency & min.\ pair frequency 2\\
Unigram & likelihood pruning & ---\\
Parity BPE \textsc{base} & cross-lingual parity & per-language streams\\
\bottomrule
\end{tabular}
\caption{The systems compared. All share the 200{,}024-identifier budget, the
byte-level pre-tokenizer, the encoder corpus and the same 41{,}347{,}986 lines
of tokenizer training text; they differ only in how the 199{,}763 learned pieces
are chosen.}
\label{tab:tokenizer-systems}
\end{table}

\paragraph{LCT.} Latent discovery reads a language-balanced sample and
estimates frequencies with a tempered estimator whose temperature interpolates
between uniform language weights ($\tau{=}0$) and raw corpus frequency
($\tau{=}1$); we evaluate $\tau \in \{0,0.2\}$. The two objectives differ in
exactly two switches on top of a shared MDL term: \textsc{MEM}
enables branching-entropy boundaries and five iterations of morphotactic
refinement, \textsc{mdl} enables neither. Remaining settings are in

\begin{table}[h]
\centering
\small
\begin{tabular}{ll}
\toprule
Component & Setting\\
\midrule
\multicolumn{2}{l}{\emph{Tokenizer}}\\
Learned tokenizer entries & 199,763 surface pieces\\
Coverage entries & 256 UTF-8 byte tokens\\
Encoder vocabulary & 200,024 entries including five specials\\
Frequency estimator & tempered, $\tau \in \{0, 0.2\}$\\
Word segmentation & script-aware (Thai-aware); protected-prefix spaces\\
Latent segmentation & 8 epochs; max piece length 20; max 200k types\\
MDL lexicon weight & 1.5\\
Morphotactics & 5 iterations\\
Surface inventory & 45\% initial base fraction; support threshold 10\\
Reconstruction & exact; round-trip accuracy 1.000\\

\bottomrule
\end{tabular}
\caption{ LCT's latent discovery.}
\label{tab:tokenizer}
\end{table}

\paragraph{BPE and Unigram.} Byte-level HuggingFace trainers on the full
corpus, with the 256-character alphabet supplied explicitly so coverage does not
depend on sampling. Both pre-tokenize with ByteLevel's regex, with the regex disabled nothing splits the text, merges span whole
100k-character chunks, and training exhausted 303\,GB without finishing.

\paragraph{Parity-aware BPE.}
Parity-aware BPE \citep{foroutan-etal-2026-parity} consumes one stream per
language and chooses merges to equalize compression across them. Every other
setting matches our BPE baseline exactly. For the held-out compression
estimates, we use the FineWeb-104 validation split.

\section{Results}

\begin{figure*}[h]
    \centering

    \includegraphics[width=\linewidth]{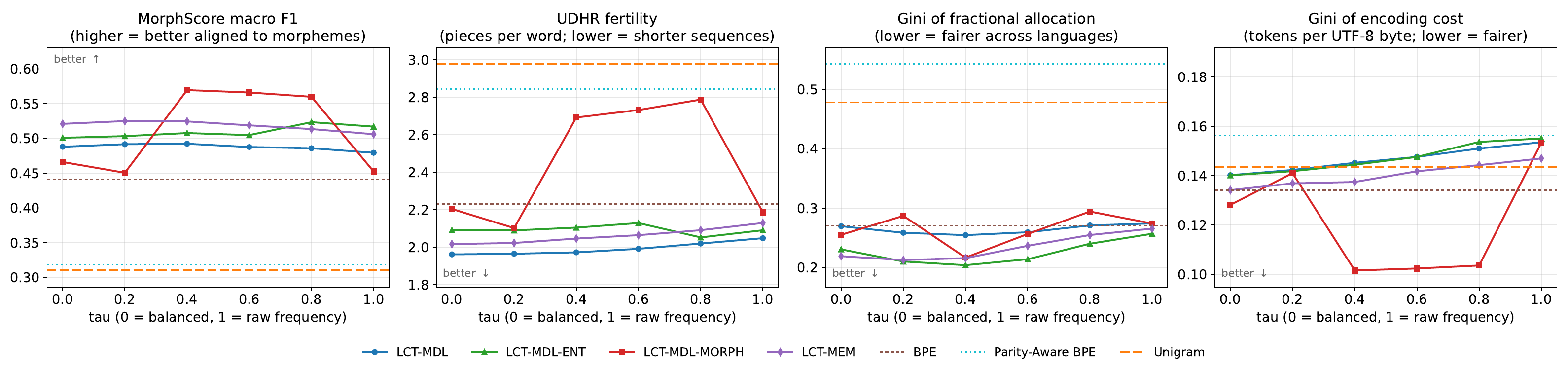}

    \vspace{0.5em}

    \includegraphics[width=\linewidth]{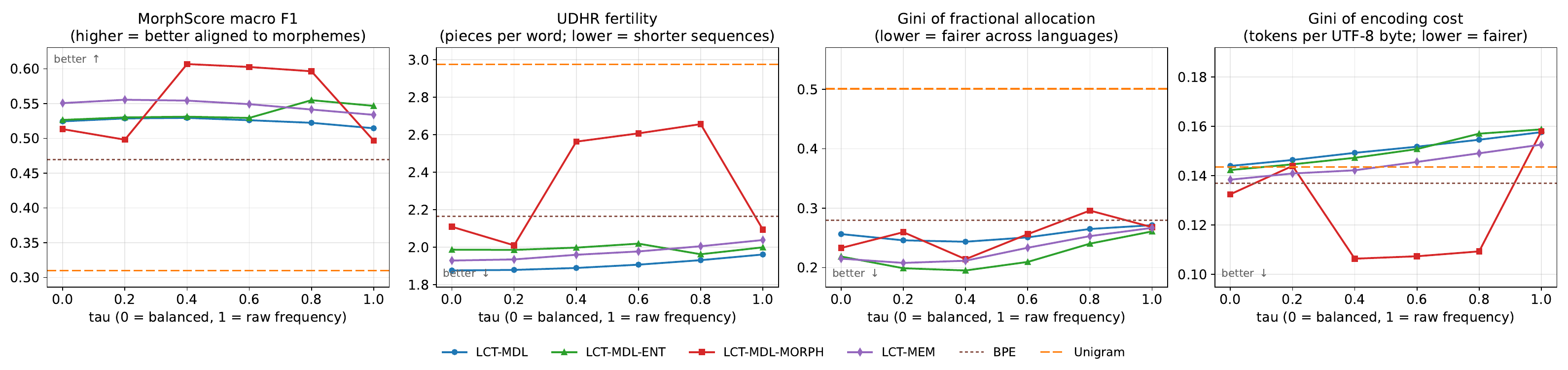}

    \caption{Intrinsic evaluation results for the 200k and 256k vocabulary configurations.}
    \label{fig:intrinsic_summary}
\end{figure*}

\clearpage
\section{Intrinsic evaluation by language}
\label{app:intrinsic-languages}

\fontsize{5}{6}\selectfont
\setlength{\tabcolsep}{3pt}
\begin{longtable}{@{}lrrrrrrrrrrrr@{}}

\toprule
 & \multicolumn{4}{c}{Fertility $\downarrow$} & \multicolumn{4}{c}{MorphScore F1 $\uparrow$} & \multicolumn{4}{c}{ Gini $\downarrow$}\\
\cmidrule(lr){2-5} \cmidrule(lr){6-9} \cmidrule(lr){10-13}
 & \shortstack{\fontsize{5}{6}\selectfont \texttt{LCT-MEM}\\$\tau=0$}  & BPE & \shortstack{Parity-Aware\\BPE} & Unigram & \shortstack{\fontsize{5}{6}\selectfont \texttt{LCT-MEM}\\$\tau=0$}  & BPE & \shortstack{Parity-Aware\\BPE} & Unigram & \shortstack{\fontsize{5}{6}\selectfont \texttt{LCT-MEM}\\$\tau=0$}  & BPE & \shortstack{Parity-Aware\\BPE} & Unigram\\
\midrule
\endfirsthead
\toprule
 & \multicolumn{4}{c}{Fertility $\downarrow$} & \multicolumn{4}{c}{MorphScore F1 $\uparrow$} & \multicolumn{4}{c}{ Gini $\downarrow$}\\
\cmidrule(lr){2-5} \cmidrule(lr){6-9} \cmidrule(lr){10-13}
 & \shortstack{\fontsize{5}{6}\selectfont \texttt{LCT-MEM}\\$\tau=0$}  & BPE & \shortstack{Parity-Aware\\BPE} & Unigram & \shortstack{\fontsize{5}{6}\selectfont \texttt{LCT-MEM}\\$\tau=0$}  & BPE & \shortstack{Parity-Aware\\BPE} & Unigram & \shortstack{\fontsize{5}{6}\selectfont \texttt{LCT-MEM}\\$\tau=0$}  & BPE & \shortstack{Parity-Aware\\BPE} & Unigram\\
\midrule
\endhead
Afrikaans & 1.431 & \textbf{1.376} & 1.482 & 1.772 & \textbf{0.397} & 0.354 & 0.270 & 0.332 & 3.67 & -0.23 & -4.26 & -2.99\\
Albanian, Tosk & 1.785 & \textbf{1.775} & 1.804 & 2.254 & \textbf{0.331} & 0.236 & 0.250 & 0.196 & 5.06 & 0.48 & -4.58 & 5.74\\
Amharic$\dagger$ & 70.610 & \textbf{50.665} & 77.457 & 62.488 &  &  &  &  & 16.89 & 1.23 & 7.45 & -0.93\\
Arabic, Standard & 1.618 & \textbf{1.513} & 1.651 & 2.056 &  &  &  &  & -4.49 & -5.60 & -5.05 & -5.00\\
Armenian & \textbf{2.002} & 2.113 & 3.035 & 3.041 & \textbf{0.412} & 0.372 & 0.319 & 0.194 & -5.02 & -5.63 & -4.80 & -5.01\\
Azerbaijani, North (Latin) & 1.761 & \textbf{1.668} & 1.983 & 2.252 & \textbf{0.496} & 0.441 & 0.346 & 0.340 & -4.16 & -4.75 & -4.95 & -2.61\\
Bamanankan & \textbf{1.834} & 1.918 & 2.470 & 2.350 &  &  &  &  & 10.58 & 13.09 & 14.45 & 9.79\\
Basque & \textbf{2.074} & 2.173 & 2.585 & 2.407 & \textbf{0.431} & 0.298 & 0.221 & 0.223 & 2.07 & 2.79 & 2.00 & 0.23\\
Belarusan & \textbf{2.267} & 2.410 & 2.824 & 2.606 & \textbf{0.349} & 0.285 & 0.180 & 0.170 & -4.13 & -5.74 & -4.74 & -5.18\\
Bhojpuri & \textbf{1.662} & 3.658 & 3.600 & 5.389 &  &  &  &  & -4.95 & 2.27 & -4.88 & 7.82\\
Breton & \textbf{2.054} & 2.231 & 2.540 & 2.480 & \textbf{0.489} & 0.443 & 0.444 & 0.418 & 11.74 & 12.67 & 6.56 & 8.11\\
Bulgarian & 1.687 & \textbf{1.677} & 2.073 & 1.945 & \textbf{0.427} & 0.359 & 0.182 & 0.204 & -4.69 & -5.86 & -5.16 & -5.43\\
Catalan-Valencian-Balear & 1.408 & \textbf{1.378} & 1.507 & 1.976 & \textbf{0.683} & 0.662 & 0.565 & 0.519 & 0.00 & -0.69 & -4.08 & 2.96\\
Chinese, Mandarin (Simplified)$\dagger$ & 30.105 & 27.629 & 38.900 & \textbf{24.474} &  &  &  &  & 7.67 & 0.00 & 1.75 & -5.08\\
Croatian & 2.040 & \textbf{2.015} & 2.482 & 2.427 & \textbf{0.407} & 0.402 & 0.218 & 0.253 & 6.81 & 1.48 & 3.43 & 3.81\\
Czech & 1.959 & \textbf{1.806} & 2.627 & 2.438 & 0.375 & \textbf{0.409} & 0.231 & 0.240 & 3.97 & -2.41 & 3.17 & 3.54\\
Danish & 1.753 & \textbf{1.669} & 2.097 & 2.076 & 0.442 & \textbf{0.484} & 0.199 & 0.344 & 0.77 & -2.61 & 0.71 & -3.98\\
Dutch & 1.419 & 1.306 & \textbf{1.189} & 1.827 & 0.467 & 0.543 & \textbf{0.796} & 0.341 & -0.50 & -3.14 & -4.95 & -3.40\\
English & 1.254 & 1.139 & \textbf{1.126} & 1.731 & 0.690 & 0.845 & \textbf{0.855} & 0.458 & -3.89 & -4.82 & -4.86 & -3.19\\
Estonian & \textbf{2.005} & 2.044 & 2.418 & 2.209 & \textbf{0.334} & 0.297 & 0.211 & 0.274 & 0.25 & 0.97 & -0.70 & -4.34\\
Finnish & 2.360 & \textbf{2.307} & 3.078 & 2.638 & \textbf{0.425} & 0.341 & 0.145 & 0.343 & -0.75 & -1.35 & 2.60 & -4.70\\
French & 1.364 & \textbf{1.300} & 1.650 & 1.938 & 0.537 & \textbf{0.609} & 0.283 & 0.303 & -3.12 & -3.33 & -1.48 & -1.83\\
Gaelic, Irish & \textbf{1.958} & 2.011 & 2.433 & 2.420 & \textbf{0.520} & 0.497 & 0.387 & 0.345 & 10.17 & 6.30 & 5.06 & 7.44\\
Gaelic, Scottish & \textbf{1.975} & 2.006 & 2.395 & 2.412 & \textbf{0.700} & 0.644 & 0.409 & 0.422 & 12.05 & 6.97 & 6.86 & 8.59\\
Galician & 1.860 & \textbf{1.606} & 2.063 & 2.452 & 0.664 & \textbf{0.707} & 0.399 & 0.374 & -2.44 & -3.50 & -2.07 & 0.96\\
Ganda & \textbf{2.373} & 2.663 & 3.089 & 2.704 &  &  &  &  & 4.24 & 11.90 & 10.93 & 4.35\\
Georgian & \textbf{2.320} & 2.576 & 4.022 & 2.736 & \textbf{0.303} & 0.141 & 0.057 & 0.091 & -4.41 & -4.87 & -5.43 & -4.40\\
German, Standard (1996) & 1.604 & \textbf{1.418} & 1.875 & 1.883 & 0.487 & \textbf{0.586} & 0.302 & 0.382 & -0.25 & -3.80 & -1.11 & -4.51\\
Greek (monotonic) & 2.261 & 1.946 & \textbf{1.939} & 2.566 & \textbf{0.757} & 0.729 & 0.729 & 0.690 & -4.36 & -5.65 & -4.93 & -5.00\\
Hausa & 1.734 & \textbf{1.614} & 1.925 & 1.998 &  &  &  &  & 11.07 & 5.63 & 3.70 & 4.63\\
Hebrew & 2.286 & \textbf{2.138} & 2.578 & 2.686 & 0.485 & \textbf{0.525} & 0.482 & 0.339 & -4.33 & -4.38 & -5.22 & -4.97\\
Hindi & \textbf{1.734} & 4.060 & 3.954 & 6.238 & \textbf{0.736} & 0.280 & 0.285 & 0.273 & -4.75 & 2.01 & -4.72 & 10.44\\
Hungarian & 1.897 & 1.803 & \textbf{1.729} & 2.310 & 0.399 & 0.444 & \textbf{0.493} & 0.379 & -1.97 & -2.95 & -4.72 & -3.61\\
Icelandic & 1.950 & 1.975 & \textbf{1.938} & 2.343 & \textbf{0.385} & 0.322 & 0.294 & 0.283 & 8.30 & 3.92 & -3.15 & 0.71\\
Igbo & \textbf{1.599} & 1.683 & 2.527 & 2.482 &  &  &  &  & 2.59 & 4.46 & 12.11 & 10.14\\
Indonesian & 1.662 & \textbf{1.493} & 1.922 & 2.086 & 0.551 & \textbf{0.714} & 0.354 & 0.281 & -4.08 & -4.90 & -3.70 & -2.82\\
Italian & 1.641 & \textbf{1.491} & 2.146 & 1.978 &  &  &  &  & -1.24 & -3.60 & -0.23 & -1.38\\
Japanese$\dagger$ & 44.867 & 37.083 & 64.067 & \textbf{31.358} &  &  &  &  & -3.50 & -5.03 & 0.47 & -5.24\\
Kazakh & \textbf{2.015} & 2.082 & 2.379 & 2.375 & \textbf{0.544} & 0.494 & 0.446 & 0.382 & -4.32 & -5.64 & -5.56 & -5.07\\
Kirghiz & 3.254 & \textbf{2.755} & 4.159 & 3.367 & \textbf{0.533} & 0.432 & 0.308 & 0.305 & -4.86 & -5.77 & -4.39 & -5.16\\
Korean & 1.870 & \textbf{1.761} & 2.363 & 2.126 & 0.829 & \textbf{0.864} & 0.706 & 0.775 & -3.68 & -4.19 & -1.69 & -5.14\\
Latvian & 2.090 & \textbf{1.848} & 2.134 & 2.462 & \textbf{0.336} & 0.321 & 0.191 & 0.156 & -1.00 & -1.98 & -2.60 & -1.15\\
Lingala & \textbf{1.431} & 1.546 & 1.910 & 1.726 &  &  &  &  & 4.79 & 6.64 & 2.86 & 2.20\\
Lithuanian & 1.880 & \textbf{1.801} & 2.523 & 2.317 & 0.407 & \textbf{0.422} & 0.108 & 0.134 & 1.80 & -0.91 & 1.49 & -2.26\\
Macedonian & \textbf{1.636} & 1.660 & 2.123 & 1.901 &  &  &  &  & -4.82 & -5.76 & -4.75 & -5.36\\
Makhuwa & \textbf{2.074} & 2.375 & 2.901 & 2.433 &  &  &  &  & 5.67 & 12.22 & 5.66 & 4.89\\
Malagasy, Plateau & \textbf{2.519} & 2.658 & 3.272 & 2.944 &  &  &  &  & 6.52 & 7.58 & 5.38 & 5.21\\
Malayalam & \textbf{4.115} & 8.889 & 8.361 & 14.745 &  &  &  &  & -4.77 & 7.26 & -2.78 & 16.53\\
Manx & \textbf{1.565} & 1.758 & 2.389 & 2.196 & \textbf{0.347} & 0.305 & 0.261 & 0.163 & 7.95 & 11.53 & 8.68 & 10.76\\
Marathi & \textbf{1.835} & 4.156 & 3.943 & 5.951 &  &  &  &  & -4.41 & 3.65 & -3.90 & 8.92\\
Norwegian, Bokmål & 1.869 & \textbf{1.760} & 2.197 & 2.022 & 0.527 & \textbf{0.574} & 0.285 & 0.425 & 1.28 & -1.76 & 0.00 & -4.16\\
Nyanja (Chechewa) & \textbf{1.689} & 1.851 & 2.572 & 2.207 &  &  &  &  & -2.89 & 3.34 & 4.12 & 1.21\\
Occitan (Auvergnat) & 2.636 & \textbf{2.622} & 2.787 & 2.915 & \textbf{0.544} & 0.452 & 0.381 & 0.436 & 16.14 & 16.03 & 9.60 & 11.07\\
Oromo, Borana-Arsi-Guji & \textbf{1.922} & 2.294 & 2.647 & 2.433 &  &  &  &  & 1.02 & 8.76 & 7.16 & 3.28\\
Pidgin, Nigerian & \textbf{1.141} & 1.159 & 1.200 & 1.591 &  &  &  &  & 2.33 & 1.73 & -3.34 & 1.69\\
Polish & 2.253 & \textbf{1.879} & 2.538 & 2.457 & 0.354 & \textbf{0.409} & 0.271 & 0.239 & 8.92 & -2.19 & 0.24 & 1.94\\
Portuguese (Brazil) & 1.760 & \textbf{1.450} & 2.062 & 2.225 & 0.488 & \textbf{0.576} & 0.190 & 0.211 & -2.21 & -4.52 & -0.88 & -1.61\\
Romanian (1953) & 1.901 & \textbf{1.624} & 2.686 & 2.201 & 0.418 & \textbf{0.453} & 0.320 & 0.279 & 5.94 & -1.13 & 10.58 & -0.47\\
Russian & 1.736 & \textbf{1.657} & 2.155 & 2.091 & \textbf{0.559} & 0.493 & 0.275 & 0.341 & -5.02 & -5.20 & -4.90 & -5.35\\
Rwanda & \textbf{1.818} & 2.021 & 2.534 & 2.515 &  &  &  &  & 4.51 & 8.17 & 7.77 & 6.03\\
Saami, North & \textbf{2.374} & 2.717 & 3.589 & 2.889 & \textbf{0.303} & 0.108 & 0.136 & 0.190 & -1.49 & 7.87 & 12.44 & -0.70\\
Sanskrit & \textbf{3.026} & 7.010 & 6.809 & 10.617 &  &  &  &  & -4.86 & 5.01 & -2.95 & 11.45\\
Serbian (Cyrillic) & \textbf{1.879} & 1.917 & 2.173 & 2.133 &  &  &  &  & -4.24 & -5.62 & -5.29 & -5.29\\
Shona & \textbf{2.344} & 2.629 & 3.431 & 2.787 &  &  &  &  & 2.86 & 8.45 & 9.95 & -0.23\\
Sinhala & \textbf{2.240} & 4.212 & 3.910 & 7.293 &  &  &  &  & -4.94 & 3.07 & -5.11 & 14.54\\
Slovak & 2.161 & \textbf{2.058} & 2.764 & 2.619 & \textbf{0.446} & 0.432 & 0.304 & 0.302 & 5.36 & -0.46 & 4.75 & 4.09\\
Slovenian & 1.749 & \textbf{1.688} & 2.294 & 2.152 & 0.400 & \textbf{0.411} & 0.232 & 0.265 & 7.08 & 0.73 & 2.26 & 0.00\\
Somali & \textbf{1.681} & 1.729 & 2.218 & 2.194 &  &  &  &  & 9.55 & 5.99 & 10.27 & 6.96\\
Sorbian, Upper & \textbf{2.061} & 2.256 & 2.559 & 2.562 & \textbf{0.389} & 0.273 & 0.221 & 0.229 & 8.63 & 10.42 & 8.38 & 6.39\\
Sotho, Southern & \textbf{1.504} & 1.673 & 1.982 & 1.791 &  &  &  &  & 12.36 & 9.99 & 6.25 & 6.68\\
Spanish & 1.675 & \textbf{1.209} & 1.901 & 2.031 & 0.472 & \textbf{0.588} & 0.173 & 0.194 & -2.67 & -4.59 & -1.88 & -2.04\\
Swahili & \textbf{1.578} & 1.707 & 2.008 & 1.909 & 0.314 & 0.271 & 0.220 & \textbf{0.368} & 3.41 & 2.53 & 1.22 & 2.46\\
Swedish & 1.815 & \textbf{1.718} & 2.124 & 2.110 & \textbf{0.455} & 0.443 & 0.265 & 0.354 & 3.12 & -1.55 & -0.47 & -4.53\\
Tamil & \textbf{2.717} & 7.503 & 7.212 & 13.287 & \textbf{0.597} & 0.259 & 0.259 & 0.157 & -4.26 & 4.19 & -3.52 & 16.05\\
Tatar & \textbf{1.832} & 2.061 & 3.285 & 2.424 & \textbf{0.575} & 0.450 & 0.276 & 0.328 & -4.47 & -5.29 & -1.28 & -5.14\\
Thai$\dagger$ & \textbf{8.187} & 13.394 & 11.318 & 18.758 &  &  &  &  & -4.79 & -5.15 & -4.89 & -5.17\\
Tigrigna & \textbf{2.391} & 2.443 & 4.858 & 2.985 &  &  &  &  & -4.58 & -3.99 & 13.00 & -4.89\\
Tswana & \textbf{1.738} & 1.849 & 2.362 & 2.076 &  &  &  &  & 9.23 & 9.52 & 8.98 & 5.48\\
Turkish & 1.852 & \textbf{1.598} & 2.341 & 2.123 & 0.432 & \textbf{0.459} & 0.310 & 0.346 & -1.73 & -2.75 & 0.96 & -3.77\\
Twi (Asante) & \textbf{1.851} & 2.028 & 2.915 & 2.702 &  &  &  &  & 11.43 & 13.97 & 16.35 & 12.90\\
Ukrainian & 1.985 & \textbf{1.948} & 2.473 & 2.304 & \textbf{0.381} & 0.369 & 0.151 & 0.263 & -4.59 & -5.77 & -5.56 & -5.39\\
Urdu & 1.912 & \textbf{1.769} & 2.067 & 2.216 & 0.538 & \textbf{0.576} & 0.512 & 0.345 & -4.32 & -5.47 & -5.42 & -5.12\\
Uyghur (Arabic) & \textbf{2.115} & 2.358 & 5.396 & 2.558 & \textbf{0.626} & 0.493 & 0.217 & 0.344 & -4.57 & -5.25 & 15.07 & -5.09\\
Uzbek, Northern (Latin) & 2.967 & \textbf{2.656} & 2.841 & 3.376 & 0.450 & \textbf{0.458} & 0.367 & 0.313 & 1.54 & 0.24 & -2.27 & 1.45\\
Veps & \textbf{2.147} & 2.426 & 2.926 & 2.688 &  &  &  &  & 7.36 & 11.16 & 4.40 & 2.70\\
Vietnamese & 1.654 & \textbf{1.243} & 1.497 & 3.075 &  &  &  &  & -4.50 & -5.15 & -4.59 & 16.95\\
Welsh & \textbf{1.854} & 1.889 & 2.386 & 2.400 & \textbf{0.672} & 0.668 & 0.612 & 0.471 & 12.84 & 9.19 & 11.69 & 11.81\\
Wolof & \textbf{1.682} & 1.913 & 2.254 & 2.196 & \textbf{0.569} & 0.326 & 0.331 & 0.210 & 13.40 & 14.69 & 9.31 & 9.39\\
Xhosa & 2.821 & \textbf{2.686} & 3.653 & 3.286 &  &  &  &  & 0.51 & 5.30 & 8.07 & 0.47\\
Yakut & \textbf{1.674} & 2.418 & 3.755 & 2.718 & \textbf{0.506} & 0.342 & 0.279 & 0.310 & -4.99 & -5.37 & -2.47 & -5.51\\
Yoruba & \textbf{2.534} & 2.936 & 3.932 & 3.880 &  &  &  &  & 6.24 & 10.83 & 13.87 & 14.97\\
Zulu & \textbf{2.028} & 2.324 & 3.227 & 2.517 &  &  &  &  & -3.33 & 4.75 & 5.96 & -2.43\\
Éwé & \textbf{2.690} & 3.329 & 4.295 & 3.656 &  &  &  &  & 9.88 & 13.50 & 18.63 & 12.42\\
\bottomrule

\caption{Intrinsic evaluation (Fertility, MorphScore and Gini) by language}
\label{tab:tokeval-lang-quality}\\

\end{longtable}


{\fontsize{5}{6}\selectfont
\setlength{\tabcolsep}{3pt}
\begin{longtable}{@{}lrrrrrrrrrrrrrrrr@{}}

\toprule
 & \multicolumn{4}{c}{Comp. Rate $\uparrow$} & \multicolumn{4}{c}{TTR $\uparrow$} & \multicolumn{4}{c}{Vocab. Util. $\uparrow$} & \multicolumn{4}{c}{R\'enyi $\uparrow$}\\
\cmidrule(lr){2-5} \cmidrule(lr){6-9} \cmidrule(lr){10-13} \cmidrule(lr){14-17}
 & \shortstack{\fontsize{5}{6}\selectfont \texttt{LCT-MEM}\\$\tau=0$}  & BPE & \shortstack{Parity-Aware\\BPE} & Unigram & \shortstack{\fontsize{5}{6}\selectfont \texttt{LCT-MEM}\\$\tau=0$}  & BPE & \shortstack{Parity-Aware\\BPE} & Unigram & \shortstack{\fontsize{5}{6}\selectfont \texttt{LCT-MEM}\\$\tau=0$}  & BPE & \shortstack{Parity-Aware\\BPE} & Unigram & \shortstack{\fontsize{5}{6}\selectfont \texttt{LCT-MEM}\\$\tau=0$}  & BPE & \shortstack{Parity-Aware\\BPE} & Unigram\\
\midrule
\endfirsthead
\toprule
 & \multicolumn{4}{c}{Comp. Rate $\uparrow$} & \multicolumn{4}{c}{TTR $\uparrow$} & \multicolumn{4}{c}{Vocab. Util. $\uparrow$} & \multicolumn{4}{c}{R\'enyi $\uparrow$}\\
\cmidrule(lr){2-5} \cmidrule(lr){6-9} \cmidrule(lr){10-13} \cmidrule(lr){14-17}
 & \shortstack{\fontsize{5}{6}\selectfont \texttt{LCT-MEM}\\$\tau=0$}  & BPE & \shortstack{Parity-Aware\\BPE} & Unigram & \shortstack{\fontsize{5}{6}\selectfont \texttt{LCT-MEM}\\$\tau=0$}  & BPE & \shortstack{Parity-Aware\\BPE} & Unigram & \shortstack{\fontsize{5}{6}\selectfont \texttt{LCT-MEM}\\$\tau=0$}  & BPE & \shortstack{Parity-Aware\\BPE} & Unigram & \shortstack{\fontsize{5}{6}\selectfont \texttt{LCT-MEM}\\$\tau=0$}  & BPE & \shortstack{Parity-Aware\\BPE} & Unigram\\
\midrule
\endhead
Afrikaans & 3.917 & \textbf{4.128} & 3.770 & 3.435 & 0.270 & \textbf{0.286} & 0.252 & 0.221 & 3.57 & \textbf{3.58} & 3.45 & 3.32 & 0.356 & 0.353 & \textbf{0.361} & 0.322\\
Albanian, Tosk & 3.858 & \textbf{3.958} & 3.826 & 2.813 & 0.260 & \textbf{0.276} & 0.265 & 0.176 & 4.25 & \textbf{4.39} & 4.35 & 3.93 & 0.328 & 0.325 & \textbf{0.329} & 0.228\\
Amharic & 2.798 & \textbf{3.899} & 2.550 & 3.161 & 0.097 & 0.108 & 0.018 & \textbf{0.136} & 2.83 & 2.25 & 0.57 & \textbf{3.50} & 0.174 & 0.197 & \textbf{0.245} & 0.178\\
Arabic, Standard & 5.487 & \textbf{5.987} & 5.354 & 4.548 & 0.370 & \textbf{0.410} & 0.327 & 0.283 & 4.63 & \textbf{4.69} & 4.19 & 4.27 & 0.407 & 0.414 & \textbf{0.418} & 0.323\\
Armenian & \textbf{7.368} & 6.982 & 4.873 & 4.837 & \textbf{0.265} & 0.230 & 0.104 & 0.179 & 4.18 & 3.82 & 2.48 & \textbf{4.28} & 0.389 & \textbf{0.398} & 0.396 & 0.256\\
Azerbaijani, North (Latin) & 4.694 & \textbf{5.038} & 3.984 & 3.373 & 0.335 & \textbf{0.377} & 0.268 & 0.243 & 4.47 & \textbf{4.69} & 4.21 & 4.50 & 0.379 & 0.374 & \textbf{0.396} & 0.307\\
Bamanankan & \textbf{3.593} & 3.074 & 2.174 & 2.474 & \textbf{0.211} & 0.177 & 0.104 & 0.124 & \textbf{2.73} & 2.68 & 2.23 & 2.33 & 0.351 & \textbf{0.360} & 0.275 & 0.270\\
Basque & \textbf{3.978} & 3.780 & 2.921 & 3.127 & 0.307 & \textbf{0.329} & 0.210 & 0.241 & 4.21 & \textbf{4.75} & 3.91 & 4.19 & 0.367 & 0.380 & \textbf{0.391} & 0.321\\
Belarusan & \textbf{6.473} & 6.014 & 5.239 & 5.245 & \textbf{0.303} & 0.277 & 0.204 & 0.246 & 5.08 & 4.99 & 4.23 & \textbf{5.08} & 0.378 & 0.387 & \textbf{0.396} & 0.319\\
Bhojpuri & \textbf{8.090} & 3.806 & 3.893 & 2.627 & \textbf{0.247} & 0.058 & 0.067 & 0.018 & \textbf{3.47} & 1.74 & 1.96 & 0.77 & \textbf{0.315} & 0.260 & 0.258 & 0.257\\
Breton & \textbf{3.498} & 3.100 & 2.561 & 2.626 & \textbf{0.266} & 0.231 & 0.167 & 0.175 & \textbf{4.18} & 4.11 & 3.60 & 3.67 & 0.381 & \textbf{0.382} & 0.380 & 0.285\\
Bulgarian & 7.011 & \textbf{7.194} & 5.375 & 5.808 & 0.282 & \textbf{0.304} & 0.190 & 0.234 & 4.14 & \textbf{4.35} & 3.64 & 4.16 & 0.343 & 0.339 & \textbf{0.369} & 0.301\\
Catalan-Valencian-Balear & 4.046 & \textbf{4.177} & 3.759 & 2.979 & 0.293 & \textbf{0.296} & 0.265 & 0.203 & \textbf{4.11} & 4.03 & 4.01 & 3.87 & 0.355 & 0.352 & \textbf{0.364} & 0.246\\
Chinese, Mandarin (Simplified) & 3.751 & 4.127 & 2.927 & \textbf{4.624} & 0.300 & 0.336 & 0.210 & \textbf{0.402} & 3.29 & 3.36 & 2.95 & \textbf{3.58} & 0.330 & 0.354 & 0.362 & \textbf{0.368}\\
Croatian & 3.772 & \textbf{3.874} & 2.767 & 2.889 & 0.327 & \textbf{0.354} & 0.212 & 0.239 & 4.36 & \textbf{4.58} & 3.84 & 4.15 & 0.378 & 0.375 & \textbf{0.399} & 0.251\\
Czech & 3.881 & \textbf{4.375} & 2.768 & 2.907 & 0.322 & \textbf{0.385} & 0.188 & 0.225 & 4.57 & \textbf{4.85} & 3.75 & 4.27 & 0.382 & 0.371 & \textbf{0.399} & 0.259\\
Danish & 4.027 & \textbf{4.404} & 3.078 & 3.502 & 0.297 & \textbf{0.337} & 0.203 & 0.237 & 4.48 & \textbf{4.65} & 4.01 & 4.12 & 0.369 & 0.360 & \textbf{0.393} & 0.275\\
Dutch & 4.082 & 4.580 & \textbf{5.311} & 3.454 & 0.263 & 0.305 & \textbf{0.320} & 0.207 & 4.08 & \textbf{4.22} & 3.82 & 3.79 & \textbf{0.354} & 0.344 & 0.325 & 0.289\\
English & 4.485 & 5.163 & \textbf{5.263} & 3.454 & 0.294 & 0.309 & \textbf{0.309} & 0.226 & 3.45 & 3.15 & 3.10 & \textbf{3.46} & \textbf{0.333} & 0.315 & 0.313 & 0.273\\
Estonian & \textbf{4.035} & 3.934 & 3.128 & 3.553 & 0.327 & \textbf{0.339} & 0.233 & 0.288 & 4.47 & \textbf{4.75} & 4.11 & 4.47 & 0.372 & 0.384 & \textbf{0.400} & 0.310\\
Finnish & 4.119 & \textbf{4.256} & 2.814 & 3.745 & 0.362 & \textbf{0.389} & 0.196 & 0.324 & 5.56 & \textbf{5.78} & 4.40 & 5.48 & 0.382 & 0.380 & \textbf{0.403} & 0.319\\
French & 4.273 & \textbf{4.607} & 3.452 & 3.205 & 0.301 & \textbf{0.305} & 0.226 & 0.210 & \textbf{4.36} & 4.09 & 4.05 & 4.05 & 0.357 & 0.342 & \textbf{0.371} & 0.299\\
Gaelic, Irish & \textbf{3.634} & 3.497 & 2.599 & 2.662 & \textbf{0.285} & 0.253 & 0.135 & 0.178 & \textbf{4.73} & 4.37 & 3.14 & 4.03 & 0.384 & \textbf{0.388} & 0.380 & 0.258\\
Gaelic, Scottish & \textbf{3.494} & 3.438 & 2.556 & 2.565 & \textbf{0.249} & 0.230 & 0.124 & 0.146 & \textbf{4.50} & 4.22 & 3.07 & 3.59 & 0.363 & \textbf{0.363} & 0.358 & 0.249\\
Galician & 4.208 & \textbf{4.657} & 3.521 & 3.053 & 0.298 & \textbf{0.330} & 0.253 & 0.212 & 4.07 & 4.06 & \textbf{4.13} & 3.97 & 0.357 & 0.349 & \textbf{0.377} & 0.269\\
Ganda & \textbf{3.880} & 3.139 & 2.472 & 2.866 & \textbf{0.389} & 0.273 & 0.165 & 0.252 & \textbf{5.17} & 4.48 & 3.45 & 4.52 & \textbf{0.400} & 0.379 & 0.377 & 0.297\\
Georgian & \textbf{10.484} & 9.048 & 5.510 & 7.677 & \textbf{0.297} & 0.234 & 0.061 & 0.240 & 4.47 & 4.09 & 1.75 & \textbf{4.93} & 0.378 & \textbf{0.393} & 0.377 & 0.281\\
German, Standard (1996) & 4.061 & \textbf{4.795} & 3.302 & 3.740 & 0.311 & \textbf{0.376} & 0.237 & 0.264 & 4.61 & \textbf{4.71} & 4.30 & 4.23 & 0.379 & 0.364 & \textbf{0.395} & 0.316\\
Greek (monotonic) & 5.884 & 6.612 & \textbf{6.661} & 4.401 & 0.251 & 0.289 & \textbf{0.291} & 0.190 & 4.81 & \textbf{4.93} & 4.93 & 4.86 & \textbf{0.395} & 0.384 & 0.383 & 0.227\\
Hausa & 3.525 & \textbf{3.573} & 2.761 & 2.857 & 0.226 & \textbf{0.244} & 0.161 & 0.188 & 3.56 & \textbf{3.79} & 3.24 & 3.64 & 0.330 & 0.331 & \textbf{0.351} & 0.285\\
Hebrew & 4.745 & \textbf{4.810} & 4.054 & 3.837 & 0.283 & \textbf{0.318} & 0.207 & 0.233 & 3.86 & \textbf{4.28} & 3.30 & 3.93 & 0.392 & 0.398 & \textbf{0.403} & 0.285\\
Hindi & \textbf{9.077} & 3.820 & 3.869 & 2.460 & \textbf{0.255} & 0.049 & 0.054 & 0.011 & \textbf{4.19} & 1.90 & 2.08 & 0.67 & \textbf{0.374} & 0.256 & 0.255 & 0.242\\
Hungarian & 4.179 & 4.548 & \textbf{4.800} & 3.460 & 0.310 & 0.345 & \textbf{0.354} & 0.244 & 4.86 & \textbf{4.96} & 4.82 & 4.62 & \textbf{0.375} & 0.367 & 0.360 & 0.264\\
Icelandic & 3.714 & 3.667 & \textbf{3.715} & 3.080 & 0.299 & 0.307 & \textbf{0.312} & 0.231 & 4.50 & 4.68 & \textbf{4.68} & 4.19 & 0.362 & \textbf{0.365} & 0.364 & 0.287\\
Igbo & \textbf{3.960} & 3.651 & 2.353 & 2.461 & \textbf{0.202} & 0.196 & 0.090 & 0.105 & 3.13 & \textbf{3.30} & 2.35 & 2.62 & \textbf{0.357} & 0.352 & 0.299 & 0.231\\
Indonesian & 4.525 & \textbf{5.280} & 3.746 & 3.382 & 0.284 & \textbf{0.316} & 0.228 & 0.207 & \textbf{3.89} & 3.71 & 3.78 & 3.80 & 0.368 & 0.354 & \textbf{0.388} & 0.291\\
Italian & 4.129 & \textbf{4.791} & 3.120 & 3.191 & 0.303 & \textbf{0.332} & 0.216 & 0.223 & \textbf{4.64} & 4.38 & 4.38 & 4.41 & 0.377 & 0.367 & \textbf{0.397} & 0.266\\
Japanese & 4.407 & 5.333 & 3.089 & \textbf{6.301} & 0.252 & 0.305 & 0.138 & \textbf{0.378} & 3.40 & 3.41 & 2.66 & \textbf{3.57} & 0.263 & 0.313 & \textbf{0.345} & 0.298\\
Kazakh & \textbf{6.649} & 6.450 & 5.521 & 5.215 & \textbf{0.316} & 0.304 & 0.210 & 0.257 & 4.79 & 4.76 & 3.85 & \textbf{4.97} & 0.391 & 0.397 & \textbf{0.406} & 0.309\\
Kirghiz & \textbf{7.189} & 6.646 & 3.825 & 5.407 & \textbf{0.311} & 0.286 & 0.096 & 0.248 & 4.66 & 4.63 & 2.71 & \textbf{4.94} & 0.394 & \textbf{0.411} & 0.346 & 0.278\\
Korean & 4.466 & \textbf{4.806} & 3.459 & 3.855 & 0.259 & \textbf{0.295} & 0.167 & 0.233 & 3.27 & \textbf{3.47} & 2.72 & 3.41 & 0.383 & 0.384 & \textbf{0.388} & 0.312\\
Latvian & 4.119 & \textbf{4.363} & 3.652 & 3.185 & 0.327 & \textbf{0.358} & 0.288 & 0.233 & 4.56 & \textbf{4.72} & 4.54 & 4.21 & 0.364 & 0.359 & \textbf{0.379} & 0.323\\
Lingala & \textbf{3.859} & 3.464 & 2.809 & 3.007 & \textbf{0.200} & 0.162 & 0.118 & 0.145 & \textbf{2.61} & 2.35 & 2.11 & 2.43 & 0.303 & 0.315 & \textbf{0.335} & 0.289\\
Lithuanian & 3.993 & \textbf{4.226} & 2.946 & 3.245 & 0.342 & \textbf{0.365} & 0.214 & 0.254 & 4.93 & \textbf{4.98} & 4.19 & 4.51 & 0.367 & 0.365 & \textbf{0.386} & 0.336\\
Macedonian & \textbf{7.026} & 6.984 & 5.076 & 5.749 & \textbf{0.302} & 0.298 & 0.179 & 0.242 & \textbf{4.20} & 4.16 & 3.44 & 4.11 & 0.355 & 0.355 & \textbf{0.382} & 0.305\\
Makhuwa & \textbf{3.807} & 3.137 & 2.579 & 2.851 & \textbf{0.327} & 0.235 & 0.163 & 0.205 & \textbf{3.96} & 3.45 & 2.92 & 3.31 & 0.380 & 0.379 & \textbf{0.385} & 0.311\\
Malagasy, Plateau & \textbf{3.781} & 3.411 & 2.580 & 2.818 & \textbf{0.273} & 0.238 & 0.141 & 0.177 & \textbf{4.44} & 4.28 & 3.35 & 3.85 & 0.344 & 0.343 & \textbf{0.359} & 0.308\\
Malayalam & \textbf{8.614} & 3.432 & 3.676 & 1.998 & \textbf{0.250} & 0.039 & 0.054 & 0.008 & \textbf{4.30} & 1.69 & 2.18 & 0.63 & \textbf{0.291} & 0.222 & 0.226 & 0.213\\
Manx & \textbf{3.747} & 3.157 & 2.525 & 2.455 & \textbf{0.231} & 0.181 & 0.123 & 0.128 & \textbf{3.57} & 3.32 & 2.83 & 3.03 & 0.368 & \textbf{0.375} & 0.374 & 0.273\\
Marathi & \textbf{10.243} & 3.674 & 3.749 & 2.551 & \textbf{0.317} & 0.044 & 0.050 & 0.010 & \textbf{4.76} & 1.83 & 2.05 & 0.61 & \textbf{0.410} & 0.230 & 0.228 & 0.233\\
Norwegian, Bokmål & 4.006 & \textbf{4.361} & 3.111 & 3.525 & 0.305 & \textbf{0.343} & 0.223 & 0.244 & 4.32 & \textbf{4.46} & 4.07 & 3.93 & 0.370 & 0.363 & \textbf{0.392} & 0.274\\
Nyanja (Chechewa) & \textbf{4.265} & 3.733 & 2.659 & 3.044 & \textbf{0.316} & 0.262 & 0.152 & 0.198 & \textbf{3.97} & 3.76 & 3.06 & 3.48 & 0.360 & 0.376 & \textbf{0.380} & 0.295\\
Occitan (Auvergnat) & \textbf{2.864} & 2.749 & 2.510 & 2.454 & \textbf{0.194} & 0.176 & 0.139 & 0.152 & \textbf{3.77} & 3.57 & 3.09 & 3.45 & \textbf{0.368} & 0.353 & 0.342 & 0.310\\
Oromo, Borana-Arsi-Guji & \textbf{4.021} & 3.392 & 2.551 & 2.909 & \textbf{0.281} & 0.249 & 0.148 & 0.196 & 3.69 & \textbf{3.88} & 3.06 & 3.56 & 0.382 & \textbf{0.388} & 0.367 & 0.310\\
Pidgin, Nigerian & \textbf{3.970} & 3.874 & 3.720 & 3.035 & \textbf{0.121} & 0.116 & 0.113 & 0.093 & 2.36 & 2.33 & 2.35 & \textbf{2.38} & 0.327 & 0.330 & \textbf{0.332} & 0.262\\
Polish & 3.684 & \textbf{4.370} & 3.095 & 3.021 & 0.304 & \textbf{0.370} & 0.234 & 0.236 & 5.02 & \textbf{5.15} & 4.60 & 4.75 & 0.398 & 0.387 & \textbf{0.411} & 0.298\\
Portuguese (Brazil) & 4.184 & \textbf{4.870} & 3.302 & 3.196 & 0.290 & \textbf{0.316} & 0.220 & 0.208 & \textbf{4.01} & 3.75 & 3.85 & 3.76 & 0.356 & 0.340 & \textbf{0.368} & 0.284\\
Romanian (1953) & 3.805 & \textbf{4.231} & 2.486 & 3.147 & 0.261 & \textbf{0.299} & 0.150 & 0.209 & 4.33 & \textbf{4.45} & 3.80 & 4.19 & \textbf{0.381} & 0.376 & 0.324 & 0.281\\
Russian & 7.576 & \textbf{8.290} & 6.553 & 6.040 & 0.333 & \textbf{0.369} & 0.286 & 0.254 & 4.75 & \textbf{4.82} & 4.71 & 4.54 & 0.369 & 0.359 & \textbf{0.383} & 0.284\\
Rwanda & \textbf{3.869} & 3.403 & 2.538 & 2.800 & \textbf{0.385} & 0.302 & 0.176 & 0.246 & \textbf{4.76} & 4.25 & 3.32 & 4.20 & 0.380 & 0.372 & \textbf{0.385} & 0.301\\
Saami, North & \textbf{4.134} & 3.407 & 2.349 & 3.154 & \textbf{0.324} & 0.256 & 0.132 & 0.235 & \textbf{5.03} & 4.82 & 3.61 & 4.79 & 0.394 & \textbf{0.412} & 0.369 & 0.312\\
Sanskrit & \textbf{8.670} & 3.631 & 3.713 & 2.435 & \textbf{0.303} & 0.047 & 0.059 & 0.012 & \textbf{4.96} & 1.83 & 2.23 & 0.67 & \textbf{0.377} & 0.239 & 0.237 & 0.231\\
Serbian (Cyrillic) & \textbf{6.535} & 6.302 & 5.382 & 5.414 & \textbf{0.331} & 0.324 & 0.234 & 0.267 & 4.33 & \textbf{4.41} & 3.73 & 4.22 & 0.374 & 0.380 & \textbf{0.392} & 0.307\\
Shona & \textbf{3.951} & 3.401 & 2.495 & 3.140 & \textbf{0.339} & 0.278 & 0.155 & 0.244 & \textbf{5.31} & 5.07 & 3.85 & 4.81 & 0.405 & \textbf{0.419} & 0.400 & 0.363\\
Sinhala & \textbf{7.277} & 3.745 & 4.005 & 2.120 & \textbf{0.216} & 0.048 & 0.072 & 0.011 & \textbf{4.37} & 1.87 & 2.64 & 0.74 & \textbf{0.384} & 0.265 & 0.257 & 0.224\\
Slovak & 3.835 & \textbf{4.128} & 2.615 & 2.874 & 0.343 & \textbf{0.379} & 0.191 & 0.238 & 4.95 & \textbf{5.07} & 4.04 & 4.58 & 0.378 & 0.373 & \textbf{0.397} & 0.265\\
Slovenian & 3.769 & \textbf{3.935} & 2.908 & 3.136 & 0.303 & \textbf{0.334} & 0.204 & 0.240 & 4.26 & \textbf{4.49} & 3.71 & 4.05 & 0.372 & 0.372 & \textbf{0.395} & 0.314\\
Somali & \textbf{3.655} & 3.519 & 2.490 & 2.744 & \textbf{0.284} & 0.283 & 0.151 & 0.203 & 4.43 & \textbf{4.58} & 3.45 & 4.23 & 0.399 & \textbf{0.406} & 0.384 & 0.282\\
Sorbian, Upper & \textbf{3.688} & 3.219 & 2.527 & 2.756 & \textbf{0.315} & 0.266 & 0.169 & 0.205 & \textbf{4.74} & 4.58 & 3.71 & 4.12 & 0.388 & \textbf{0.400} & 0.377 & 0.315\\
Sotho, Southern & \textbf{3.490} & 3.261 & 2.573 & 2.746 & \textbf{0.217} & 0.195 & 0.128 & 0.154 & \textbf{3.51} & 3.37 & 2.81 & 3.15 & 0.323 & 0.330 & \textbf{0.337} & 0.272\\
Spanish & 4.232 & \textbf{5.009} & 3.490 & 3.239 & 0.277 & \textbf{0.318} & 0.236 & 0.209 & 3.95 & 3.83 & \textbf{4.08} & 3.89 & 0.344 & 0.324 & \textbf{0.363} & 0.270\\
Swahili & \textbf{3.918} & 3.781 & 2.987 & 2.987 & \textbf{0.299} & 0.289 & 0.208 & 0.214 & 3.95 & \textbf{3.95} & 3.61 & 3.71 & 0.348 & 0.352 & \textbf{0.374} & 0.285\\
Swedish & 3.949 & \textbf{4.323} & 3.125 & 3.567 & 0.296 & \textbf{0.337} & 0.215 & 0.244 & 4.53 & \textbf{4.72} & 4.17 & 4.14 & 0.378 & 0.373 & \textbf{0.398} & 0.296\\
Tamil & \textbf{11.083} & 3.655 & 3.739 & 2.013 & \textbf{0.280} & 0.034 & 0.041 & 0.006 & \textbf{4.80} & 1.74 & 2.07 & 0.58 & \textbf{0.411} & 0.214 & 0.212 & 0.167\\
Tatar & \textbf{6.668} & 5.799 & 3.431 & 5.001 & \textbf{0.308} & 0.259 & 0.090 & 0.234 & 4.36 & 4.22 & 2.47 & \textbf{4.41} & 0.380 & \textbf{0.390} & 0.294 & 0.327\\
Thai & \textbf{9.218} & 5.771 & 6.865 & 4.114 & 0.211 & 0.159 & \textbf{0.265} & 0.135 & 3.08 & 3.72 & \textbf{5.20} & 4.42 & \textbf{0.391} & 0.322 & 0.301 & 0.223\\
Tigrigna & \textbf{4.938} & 4.804 & 2.303 & 3.749 & \textbf{0.225} & 0.167 & 0.021 & 0.172 & 3.92 & 2.98 & 0.79 & \textbf{3.95} & 0.323 & \textbf{0.367} & 0.309 & 0.278\\
Tswana & \textbf{3.672} & 3.322 & 2.522 & 2.815 & \textbf{0.216} & 0.178 & 0.113 & 0.148 & \textbf{3.61} & 3.30 & 2.75 & 3.24 & 0.335 & \textbf{0.346} & 0.344 & 0.280\\
Turkish & 4.157 & \textbf{4.525} & 3.039 & 3.502 & 0.349 & \textbf{0.404} & 0.224 & 0.312 & 4.62 & \textbf{4.91} & 4.06 & 4.90 & 0.387 & 0.381 & \textbf{0.410} & 0.325\\
Twi (Asante) & \textbf{3.503} & 3.018 & 2.010 & 2.333 & \textbf{0.198} & 0.162 & 0.084 & 0.105 & \textbf{3.14} & 2.98 & 2.31 & 2.51 & 0.352 & \textbf{0.357} & 0.271 & 0.245\\
Ukrainian & 6.707 & \textbf{7.143} & 5.649 & 5.445 & 0.335 & \textbf{0.346} & 0.246 & 0.254 & \textbf{4.85} & 4.70 & 4.23 & 4.53 & 0.377 & 0.370 & \textbf{0.390} & 0.291\\
Urdu & 5.462 & \textbf{5.960} & 5.393 & 4.015 & 0.239 & \textbf{0.263} & 0.224 & 0.157 & 3.91 & \textbf{3.94} & 3.71 & 3.50 & 0.365 & 0.368 & \textbf{0.380} & 0.273\\
Uyghur (Arabic) & \textbf{6.961} & 5.477 & 2.133 & 4.907 & \textbf{0.283} & 0.178 & 0.027 & 0.176 & \textbf{4.53} & 3.62 & 1.40 & 4.00 & 0.399 & \textbf{0.414} & 0.232 & 0.401\\
Uzbek, Northern (Latin) & \textbf{4.005} & 3.991 & 3.539 & 3.040 & \textbf{0.292} & 0.290 & 0.243 & 0.217 & \textbf{4.53} & 4.51 & 4.28 & 4.44 & 0.388 & 0.395 & \textbf{0.403} & 0.330\\
Veps & \textbf{3.766} & 3.175 & 2.655 & 2.986 & \textbf{0.278} & 0.221 & 0.165 & 0.213 & \textbf{3.99} & 3.76 & 3.36 & 3.84 & 0.367 & \textbf{0.380} & 0.379 & 0.306\\
Vietnamese & 4.791 & \textbf{5.396} & 4.778 & 1.992 & 0.199 & \textbf{0.224} & 0.199 & 0.045 & \textbf{3.12} & 3.12 & 3.12 & 1.68 & 0.388 & 0.378 & \textbf{0.389} & 0.275\\
Welsh & \textbf{3.440} & 3.334 & 2.374 & 2.423 & \textbf{0.291} & 0.267 & 0.150 & 0.172 & \textbf{4.31} & 4.09 & 3.22 & 3.62 & \textbf{0.385} & 0.376 & 0.385 & 0.272\\
Wolof & \textbf{3.372} & 2.934 & 2.511 & 2.501 & \textbf{0.248} & 0.213 & 0.155 & 0.161 & \textbf{3.63} & 3.59 & 3.04 & 3.19 & 0.360 & \textbf{0.368} & 0.368 & 0.308\\
Xhosa & \textbf{4.034} & 3.617 & 2.533 & 3.117 & \textbf{0.390} & 0.326 & 0.164 & 0.262 & \textbf{5.26} & 4.90 & 3.53 & 4.57 & 0.405 & \textbf{0.419} & 0.407 & 0.340\\
Yakut & \textbf{7.824} & 5.887 & 3.544 & 5.462 & \textbf{0.345} & 0.225 & 0.077 & 0.247 & 4.65 & 4.04 & 2.28 & \textbf{4.76} & 0.383 & \textbf{0.405} & 0.359 & 0.332\\
Yoruba & \textbf{3.785} & 3.185 & 2.211 & 2.109 & \textbf{0.112} & 0.084 & 0.041 & 0.040 & \textbf{2.67} & 2.39 & 1.69 & 1.72 & \textbf{0.383} & 0.322 & 0.308 & 0.243\\
Zulu & \textbf{4.319} & 3.631 & 2.573 & 3.329 & \textbf{0.456} & 0.347 & 0.181 & 0.304 & \textbf{5.37} & 4.86 & 3.58 & 4.65 & 0.398 & \textbf{0.409} & 0.401 & 0.350\\
Éwé & \textbf{3.638} & 3.051 & 1.803 & 2.363 & \textbf{0.197} & 0.164 & 0.073 & 0.094 & \textbf{3.65} & 3.62 & 2.73 & 2.69 & 0.386 & \textbf{0.397} & 0.260 & 0.263\\
\bottomrule

\caption{Intrinsic evaluation (Comp. Rate, TTR, Vocab. Util. and R\'enyi) by language}

\label{tab:tokeval-lang-efficiency}\\
\end{longtable}
}

\normalsize



\section{Per-language transfer}

\definecolor{ULCTgain}{HTML}{1B7837}
\definecolor{ULCTloss}{HTML}{B2182B}
\providecommand{\faintrule}{%
  \arrayrulecolor{black!18}\midrule\arrayrulecolor{black}}
\tiny
\begin{longtable}{lrrrrrrrrrrr}
\toprule
& \multicolumn{7}{c}{Accuracy} & \multicolumn{4}{c}{$\Delta$ vs BPE}\\
\cmidrule(lr){2-8}\cmidrule(lr){9-12}
Language & \shortstack{\texttt{LCT-MDL}\\$\tau=0$} & \shortstack{\texttt{LCT-MDL}\\$\tau=0.2$} & \shortstack{\texttt{LCT-MEM}\\$\tau=0$} & \shortstack{\texttt{LCT-MEM}\\$\tau=0.2$} & BPE & Unigram & \shortstack{Parity-Aware\\BPE} & \shortstack{\texttt{LCT-MDL}\\$\tau=0$} & \shortstack{\texttt{LCT-MDL}\\$\tau=0.2$} & \shortstack{\texttt{LCT-MEM}\\$\tau=0$} & \shortstack{\texttt{LCT-MEM}\\$\tau=0.2$}\\
\midrule
\endfirsthead
\toprule
& \multicolumn{7}{c}{Accuracy} & \multicolumn{4}{c}{$\Delta$ vs BPE}\\
\cmidrule(lr){2-8}\cmidrule(lr){9-12}
Language & \shortstack{\texttt{LCT-MDL}\\$\tau=0$} & \shortstack{\texttt{LCT-MDL}\\$\tau=0.2$} & \shortstack{\texttt{LCT-MEM}\\$\tau=0$} & \shortstack{\texttt{LCT-MEM}\\$\tau=0.2$} & BPE & Unigram & \shortstack{Parity-Aware\\BPE} & \shortstack{\texttt{LCT-MDL}\\$\tau=0$} & \shortstack{\texttt{LCT-MDL}\\$\tau=0.2$} & \shortstack{\texttt{LCT-MEM}\\$\tau=0$} & \shortstack{\texttt{LCT-MEM}\\$\tau=0.2$}\\
\midrule
\endhead
\midrule
\multicolumn{12}{r}{\textit{continued}}\\
\endfoot
\endlastfoot
\multicolumn{12}{l}{\textbf{XNLI}, \textit{all 15 languages, grouped by family}}\\
\multicolumn{12}{l}{\textit{Indo-European} (9)}\\
\quad bulgarian & 48.97 & 50.48 & 49.56 & 47.90 & 48.89 & 47.28 & 46.02 & $+0.08$ & $+1.59$ & $+0.67$ & $-0.99$\\
\quad english & 68.14 & 67.64 & 68.07 & 68.34 & 68.59 & 69.30 & 67.39 & $-0.45$ & \cellcolor{ULCTloss!18}$-0.96$ & $-0.53$ & $-0.26$\\
\quad french & 50.18 & 49.82 & 50.06 & 48.75 & 53.64 & 46.61 & 48.63 & \cellcolor{ULCTloss!35}$-3.47$ & \cellcolor{ULCTloss!37}$-3.82$ & \cellcolor{ULCTloss!36}$-3.58$ & \cellcolor{ULCTloss!44}$-4.89$\\
\quad german & 47.99 & 47.75 & 48.83 & 47.93 & 50.32 & 44.66 & 45.90 & \cellcolor{ULCTloss!27}$-2.33$ & \cellcolor{ULCTloss!29}$-2.57$ & $-1.50$ & \cellcolor{ULCTloss!28}$-2.40$\\
\quad greek & 45.96 & 46.08 & 44.63 & 45.56 & 46.36 & 43.56 & 45.58 & $-0.39$ & $-0.27$ & $-1.72$ & $-0.79$\\
\quad hindi & 44.23 & 43.18 & 44.48 & 44.83 & 39.09 & 38.06 & 38.29 & \cellcolor{ULCTgain!45}$+5.14$ & \cellcolor{ULCTgain!39}$+4.09$ & \cellcolor{ULCTgain!45}$+5.39$ & \cellcolor{ULCTgain!45}$+5.74$\\
\quad russian & 46.32 & 48.26 & 48.57 & 46.89 & 48.55 & 45.49 & 45.09 & \cellcolor{ULCTloss!27}$-2.23$ & $-0.29$ & $+0.03$ & $-1.65$\\
\quad spanish & 55.93 & 56.69 & 56.40 & 56.61 & 56.05 & 52.68 & 54.08 & $-0.12$ & $+0.64$ & $+0.35$ & $+0.57$\\
\quad urdu & 42.85 & 43.97 & 42.48 & 42.72 & 42.93 & 40.11 & 43.82 & $-0.08$ & $+1.04$ & $-0.45$ & $-0.21$\\
\faintrule
\multicolumn{12}{l}{\textit{Atlantic-Congo} (1)}\\
\quad swahili & 45.11 & 45.01 & 43.67 & 45.10 & 44.93 & 44.68 & 44.80 & $+0.19$ & $+0.08$ & $-1.25$ & $+0.18$\\
\faintrule
\multicolumn{12}{l}{\textit{Turkic} (1)}\\
\quad turkish & 47.69 & 47.40 & 47.39 & 47.47 & 48.31 & 45.02 & 44.84 & $-0.62$ & $-0.91$ & $-0.93$ & $-0.84$\\
\faintrule
\multicolumn{12}{l}{\textit{Afro-Asiatic} (1)}\\
\quad arabic & 44.32 & 44.54 & 45.17 & 43.40 & 45.60 & 42.89 & 44.77 & $-1.27$ & $-1.05$ & $-0.43$ & \cellcolor{ULCTloss!27}$-2.20$\\
\faintrule
\multicolumn{12}{l}{\textit{Austroasiatic} (1)}\\
\quad vietnamese & 46.75 & 48.56 & 48.11 & 48.00 & 50.53 & 45.64 & 46.86 & \cellcolor{ULCTloss!37}$-3.78$ & \cellcolor{ULCTloss!25}$-1.97$ & \cellcolor{ULCTloss!28}$-2.42$ & \cellcolor{ULCTloss!29}$-2.53$\\
\faintrule
\multicolumn{12}{l}{\textit{Kra-Dai} (1)}\\
\quad thai & 45.37 & 45.02 & 46.73 & 44.75 & 37.56 & 38.59 & 36.69 & \cellcolor{ULCTgain!45}$+7.81$ & \cellcolor{ULCTgain!45}$+7.46$ & \cellcolor{ULCTgain!45}$+9.18$ & \cellcolor{ULCTgain!45}$+7.19$\\
\faintrule
\multicolumn{12}{l}{\textit{Sino-Tibetan} (1)}\\
\quad chinese & 46.13 & 45.42 & 46.06 & 45.19 & 44.47 & 40.04 & 44.20 & \cellcolor{ULCTgain!23}$+1.66$ & $+0.95$ & \cellcolor{ULCTgain!23}$+1.59$ & $+0.72$\\
\midrule
\multicolumn{12}{l}{\textbf{Belebele}, \textit{all 79 languages, grouped by family}}\\
\multicolumn{12}{l}{\textit{Indo-European} (36)}\\
\quad afrikaans & 31.42 & 31.93 & 29.84 & 32.20 & 29.93 & 30.04 & 29.67 & \cellcolor{ULCTgain!22}$+1.49$ & \cellcolor{ULCTgain!25}$+2.00$ & $-0.09$ & \cellcolor{ULCTgain!27}$+2.27$\\
\quad albanian & 30.64 & 31.58 & 30.69 & 31.44 & 30.33 & 28.87 & 30.73 & $+0.31$ & $+1.24$ & $+0.36$ & \cellcolor{ULCTgain!19}$+1.11$\\
\quad armenian & 27.76 & 29.31 & 28.02 & 29.27 & 26.93 & 27.36 & 28.11 & $+0.82$ & $+2.38$ & $+1.09$ & $+2.33$\\
\quad bulgarian & 29.58 & 30.02 & 29.80 & 32.47 & 29.07 & 28.40 & 31.96 & $+0.51$ & $+0.96$ & $+0.73$ & \cellcolor{ULCTgain!34}$+3.40$\\
\quad catalan & 30.82 & 34.58 & 32.11 & 35.60 & 32.89 & 33.67 & 31.27 & \cellcolor{ULCTloss!26}$-2.07$ & \cellcolor{ULCTgain!23}$+1.69$ & $-0.78$ & \cellcolor{ULCTgain!30}$+2.71$\\
\quad croatian & 30.40 & 31.96 & 31.09 & 32.84 & 31.71 & 32.44 & 32.91 & $-1.31$ & $+0.24$ & $-0.62$ & $+1.13$\\
\quad czech & 29.91 & 32.22 & 30.87 & 32.60 & 31.82 & 31.11 & 30.73 & \cellcolor{ULCTloss!25}$-1.91$ & $+0.40$ & $-0.96$ & $+0.78$\\
\quad danish & 30.80 & 33.04 & 31.36 & 33.71 & 31.16 & 32.36 & 31.38 & $-0.36$ & \cellcolor{ULCTgain!24}$+1.89$ & $+0.20$ & \cellcolor{ULCTgain!29}$+2.56$\\
\quad dutch & 31.69 & 32.33 & 32.04 & 33.91 & 33.56 & 33.20 & 31.51 & \cellcolor{ULCTloss!24}$-1.87$ & \cellcolor{ULCTloss!20}$-1.22$ & $-1.51$ & $+0.36$\\
\quad english & 31.38 & 34.02 & 31.33 & 34.18 & 31.62 & 31.09 & 31.04 & $-0.24$ & $+2.40$ & $-0.29$ & \cellcolor{ULCTgain!29}$+2.56$\\
\quad french & 31.07 & 32.56 & 30.56 & 32.84 & 33.04 & 32.87 & 31.73 & \cellcolor{ULCTloss!25}$-1.98$ & $-0.49$ & \cellcolor{ULCTloss!28}$-2.49$ & $-0.20$\\
\quad german & 30.87 & 34.42 & 32.29 & 34.76 & 33.47 & 32.69 & 30.69 & \cellcolor{ULCTloss!29}$-2.60$ & $+0.96$ & \cellcolor{ULCTloss!20}$-1.18$ & \cellcolor{ULCTgain!21}$+1.29$\\
\quad greek & 30.13 & 31.98 & 29.16 & 30.56 & 29.33 & 28.64 & 29.56 & $+0.80$ & \cellcolor{ULCTgain!29}$+2.64$ & $-0.18$ & $+1.22$\\
\quad hindi & 28.82 & 29.53 & 27.96 & 30.73 & 25.96 & 27.16 & 25.60 & $+2.87$ & \cellcolor{ULCTgain!36}$+3.58$ & \cellcolor{ULCTgain!25}$+2.00$ & \cellcolor{ULCTgain!44}$+4.78$\\
\quad icelandic & 31.33 & 31.04 & 31.69 & 32.40 & 31.76 & 30.89 & 31.69 & $-0.42$ & $-0.71$ & $-0.07$ & $+0.64$\\
\quad italian & 31.84 & 31.40 & 31.51 & 32.98 & 31.18 & 31.89 & 30.73 & $+0.67$ & $+0.22$ & $+0.33$ & \cellcolor{ULCTgain!24}$+1.80$\\
\quad latvian & 29.22 & 29.82 & 29.71 & 31.24 & 30.60 & 29.47 & 31.33 & $-1.38$ & $-0.78$ & $-0.89$ & $+0.64$\\
\quad lithuanian & 31.98 & 32.89 & 29.93 & 33.07 & 32.02 & 31.98 & 32.40 & $-0.04$ & $+0.87$ & \cellcolor{ULCTloss!26}$-2.09$ & $+1.04$\\
\quad macedonian & 28.67 & 28.56 & 29.20 & 30.00 & 29.04 & 28.13 & 27.58 & $-0.38$ & $-0.49$ & $+0.16$ & $+0.96$\\
\quad marathi & 30.27 & 32.31 & 31.93 & 33.36 & 24.73 & 26.47 & 25.87 & \cellcolor{ULCTgain!45}$+5.53$ & \cellcolor{ULCTgain!45}$+7.58$ & \cellcolor{ULCTgain!45}$+7.20$ & \cellcolor{ULCTgain!45}$+8.62$\\
\quad norwegian\_bokmal & 31.24 & 32.69 & 31.16 & 34.18 & 31.53 & 31.02 & 30.93 & $-0.29$ & $+1.16$ & $-0.38$ & \cellcolor{ULCTgain!29}$+2.64$\\
\quad pashto & 27.82 & 28.02 & 27.22 & 29.31 & 28.38 & 25.56 & 29.00 & \cellcolor{ULCTloss!16}$-0.56$ & $-0.36$ & $-1.16$ & $+0.93$\\
\quad persian & 31.04 & 34.02 & 31.24 & 32.78 & 32.40 & 32.64 & 29.47 & \cellcolor{ULCTloss!21}$-1.36$ & $+1.62$ & $-1.16$ & $+0.38$\\
\quad polish & 32.18 & 31.69 & 30.82 & 32.87 & 31.29 & 31.16 & 30.27 & $+0.89$ & $+0.40$ & $-0.47$ & \cellcolor{ULCTgain!22}$+1.58$\\
\quad portuguese & 30.89 & 31.27 & 30.76 & 34.02 & 32.42 & 31.69 & 30.36 & $-1.53$ & \cellcolor{ULCTloss!20}$-1.16$ & \cellcolor{ULCTloss!23}$-1.67$ & \cellcolor{ULCTgain!23}$+1.60$\\
\quad romanian & 31.44 & 32.69 & 30.69 & 32.93 & 30.60 & 31.13 & 29.84 & $+0.84$ & \cellcolor{ULCTgain!26}$+2.09$ & $+0.09$ & \cellcolor{ULCTgain!27}$+2.33$\\
\quad russian & 29.71 & 30.27 & 29.73 & 32.09 & 30.78 & 30.53 & 29.11 & \cellcolor{ULCTloss!19}$-1.07$ & $-0.51$ & \cellcolor{ULCTloss!19}$-1.04$ & \cellcolor{ULCTgain!21}$+1.31$\\
\quad serbian & 32.29 & 32.11 & 29.67 & 32.67 & 30.91 & 31.29 & 30.22 & $+1.38$ & $+1.20$ & $-1.24$ & \cellcolor{ULCTgain!24}$+1.76$\\
\quad sindhi & 28.24 & 28.51 & 28.44 & 28.16 & 30.40 & 28.42 & 26.40 & \cellcolor{ULCTloss!26}$-2.16$ & \cellcolor{ULCTloss!24}$-1.89$ & \cellcolor{ULCTloss!25}$-1.96$ & \cellcolor{ULCTloss!27}$-2.24$\\
\quad sinhala & 30.93 & 32.20 & 30.84 & 33.62 & 25.24 & 26.73 & 25.18 & \cellcolor{ULCTgain!45}$+5.69$ & \cellcolor{ULCTgain!45}$+6.96$ & \cellcolor{ULCTgain!45}$+5.60$ & \cellcolor{ULCTgain!45}$+8.38$\\
\quad slovak & 29.67 & 32.64 & 31.29 & 34.31 & 32.40 & 31.96 & 33.20 & \cellcolor{ULCTloss!30}$-2.73$ & $+0.24$ & $-1.11$ & \cellcolor{ULCTgain!25}$+1.91$\\
\quad slovenian & 31.89 & 32.58 & 30.40 & 31.58 & 30.64 & 30.27 & 29.24 & $+1.24$ & \cellcolor{ULCTgain!25}$+1.93$ & $-0.24$ & $+0.93$\\
\quad spanish & 30.80 & 31.71 & 30.84 & 33.31 & 32.89 & 30.27 & 31.62 & \cellcolor{ULCTloss!26}$-2.09$ & $-1.18$ & $-2.04$ & $+0.42$\\
\quad swedish & 31.69 & 34.09 & 33.73 & 34.78 & 32.29 & 32.40 & 30.91 & $-0.60$ & \cellcolor{ULCTgain!24}$+1.80$ & $+1.44$ & \cellcolor{ULCTgain!28}$+2.49$\\
\quad ukrainian & 30.51 & 30.71 & 30.20 & 32.40 & 31.02 & 29.11 & 29.98 & $-0.51$ & $-0.31$ & $-0.82$ & \cellcolor{ULCTgain!21}$+1.38$\\
\quad urdu & 31.67 & 31.58 & 30.40 & 31.87 & 30.07 & 29.71 & 28.98 & \cellcolor{ULCTgain!23}$+1.60$ & \cellcolor{ULCTgain!22}$+1.51$ & $+0.33$ & $+1.80$\\
\faintrule
\multicolumn{12}{l}{\textit{Atlantic-Congo} (13)}\\
\quad chichewa\_nyanja & 27.71 & 26.49 & 27.36 & 26.53 & 26.64 & 27.09 & 26.49 & \cellcolor{ULCTgain!19}$+1.07$ & $-0.16$ & $+0.71$ & $-0.11$\\
\quad igbo & 24.49 & 23.49 & 24.96 & 25.71 & 25.62 & 25.53 & 26.24 & $-1.13$ & \cellcolor{ULCTloss!26}$-2.13$ & $-0.67$ & $+0.09$\\
\quad kinyarwanda & 26.62 & 26.82 & 27.73 & 25.56 & 28.38 & 27.96 & 27.20 & \cellcolor{ULCTloss!24}$-1.76$ & \cellcolor{ULCTloss!22}$-1.56$ & $-0.64$ & \cellcolor{ULCTloss!31}$-2.82$\\
\quad lingala & 26.29 & 26.22 & 25.49 & 27.22 & 27.93 & 27.44 & 27.69 & \cellcolor{ULCTloss!23}$-1.64$ & \cellcolor{ULCTloss!23}$-1.71$ & \cellcolor{ULCTloss!28}$-2.44$ & $-0.71$\\
\quad luganda & 24.33 & 24.24 & 23.16 & 25.04 & 23.89 & 25.53 & 25.53 & $+0.44$ & $+0.36$ & $-0.73$ & $+1.16$\\
\quad sesotho & 25.40 & 27.51 & 26.93 & 27.07 & 26.78 & 27.24 & 27.11 & \cellcolor{ULCTloss!21}$-1.38$ & $+0.73$ & $+0.16$ & $+0.29$\\
\quad setswana & 24.24 & 26.07 & 26.58 & 25.67 & 27.78 & 27.16 & 26.47 & \cellcolor{ULCTloss!35}$-3.53$ & \cellcolor{ULCTloss!23}$-1.71$ & $-1.20$ & \cellcolor{ULCTloss!26}$-2.11$\\
\quad shona & 28.09 & 30.00 & 28.96 & 29.98 & 29.33 & 28.80 & 28.42 & $-1.24$ & $+0.67$ & $-0.38$ & $+0.64$\\
\quad swahili & 28.24 & 28.16 & 27.44 & 29.31 & 28.80 & 28.87 & 28.69 & $-0.56$ & \cellcolor{ULCTloss!16}$-0.64$ & $-1.36$ & $+0.51$\\
\quad wolof & 26.38 & 25.16 & 25.82 & 26.80 & 25.93 & 24.93 & 24.53 & $+0.44$ & $-0.78$ & $-0.11$ & \cellcolor{ULCTgain!18}$+0.87$\\
\quad xhosa & 27.04 & 25.40 & 26.84 & 28.16 & 27.82 & 26.93 & 25.09 & $-0.78$ & \cellcolor{ULCTloss!28}$-2.42$ & $-0.98$ & $+0.33$\\
\quad yoruba & 25.04 & 27.18 & 25.76 & 26.47 & 24.13 & 24.82 & 24.67 & $+0.91$ & \cellcolor{ULCTgain!32}$+3.04$ & $+1.62$ & \cellcolor{ULCTgain!27}$+2.33$\\
\quad zulu & 26.71 & 26.64 & 26.29 & 26.78 & 27.44 & 27.51 & 28.82 & $-0.73$ & $-0.80$ & $-1.16$ & $-0.67$\\
\faintrule
\multicolumn{12}{l}{\textit{Turkic} (5)}\\
\quad azerbaijani & 26.78 & 28.24 & 28.71 & 28.93 & 26.87 & 28.60 & 27.82 & $-0.09$ & $+1.38$ & \cellcolor{ULCTgain!24}$+1.84$ & $+2.07$\\
\quad kazakh & 28.87 & 29.62 & 29.56 & 29.36 & 28.22 & 28.27 & 30.09 & $+0.64$ & $+1.40$ & $+1.33$ & $+1.13$\\
\quad kyrgyz & 28.60 & 29.67 & 29.98 & 29.42 & 29.02 & 29.67 & 29.22 & $-0.42$ & $+0.64$ & $+0.96$ & $+0.40$\\
\quad turkish & 28.33 & 30.56 & 30.73 & 33.67 & 32.00 & 29.02 & 30.24 & \cellcolor{ULCTloss!36}$-3.67$ & $-1.44$ & \cellcolor{ULCTloss!20}$-1.27$ & $+1.67$\\
\quad uzbek & 29.04 & 30.67 & 29.73 & 31.98 & 29.42 & 28.38 & 29.96 & $-0.38$ & $+1.24$ & $+0.31$ & $+2.56$\\
\faintrule
\multicolumn{12}{l}{\textit{Uralic} (3)}\\
\quad estonian & 29.07 & 31.53 & 30.84 & 32.58 & 29.44 & 30.31 & 29.73 & $-0.38$ & \cellcolor{ULCTgain!26}$+2.09$ & $+1.40$ & \cellcolor{ULCTgain!33}$+3.13$\\
\quad finnish & 30.71 & 33.82 & 32.07 & 35.33 & 32.33 & 31.80 & 30.78 & $-1.62$ & $+1.49$ & $-0.27$ & \cellcolor{ULCTgain!32}$+3.00$\\
\quad hungarian & 31.67 & 31.20 & 31.11 & 31.51 & 28.87 & 31.07 & 29.11 & \cellcolor{ULCTgain!30}$+2.80$ & \cellcolor{ULCTgain!27}$+2.33$ & \cellcolor{ULCTgain!27}$+2.24$ & $+2.64$\\
\faintrule
\multicolumn{12}{l}{\textit{Afro-Asiatic} (7)}\\
\quad amharic & 26.71 & 28.09 & 28.44 & 30.56 & 27.00 & 26.76 & 27.09 & $-0.29$ & $+1.09$ & $+1.44$ & \cellcolor{ULCTgain!35}$+3.56$\\
\quad arabic & 29.38 & 30.91 & 29.04 & 31.53 & 31.29 & 31.36 & 29.04 & \cellcolor{ULCTloss!25}$-1.91$ & $-0.38$ & \cellcolor{ULCTloss!27}$-2.24$ & $+0.24$\\
\quad hausa & 27.96 & 27.02 & 26.96 & 25.96 & 27.64 & 27.22 & 26.36 & $+0.31$ & $-0.62$ & $-0.69$ & \cellcolor{ULCTloss!23}$-1.69$\\
\quad hebrew & 30.58 & 31.16 & 32.53 & 33.36 & 32.49 & 27.93 & 31.02 & \cellcolor{ULCTloss!25}$-1.91$ & \cellcolor{ULCTloss!21}$-1.33$ & $+0.04$ & $+0.87$\\
\quad oromo & 26.76 & 28.18 & 27.11 & 28.82 & 27.69 & 25.84 & 25.87 & \cellcolor{ULCTloss!18}$-0.93$ & $+0.49$ & $-0.58$ & $+1.13$\\
\quad somali & 27.82 & 27.31 & 25.91 & 28.02 & 25.53 & 25.47 & 26.20 & \cellcolor{ULCTgain!27}$+2.29$ & \cellcolor{ULCTgain!24}$+1.78$ & $+0.38$ & \cellcolor{ULCTgain!28}$+2.49$\\
\quad tigrinya & 28.98 & 28.13 & 30.56 & 29.42 & 30.27 & 28.33 & 27.60 & $-1.29$ & \cellcolor{ULCTloss!26}$-2.13$ & $+0.29$ & $-0.84$\\
\faintrule
\multicolumn{12}{l}{\textit{Austronesian} (4)}\\
\quad filipino & 27.24 & 29.93 & 28.60 & 29.93 & 30.07 & 29.18 & 30.42 & \cellcolor{ULCTloss!31}$-2.82$ & $-0.13$ & $-1.47$ & $-0.13$\\
\quad indonesian & 30.98 & 31.18 & 30.60 & 32.20 & 31.31 & 29.78 & 31.00 & $-0.33$ & $-0.13$ & $-0.71$ & $+0.89$\\
\quad malagasy & 26.87 & 28.76 & 28.98 & 29.62 & 29.13 & 28.62 & 29.93 & $-2.27$ & $-0.38$ & $-0.16$ & $+0.49$\\
\quad malay & 30.78 & 31.00 & 30.69 & 33.82 & 30.58 & 30.27 & 30.91 & $+0.20$ & $+0.42$ & $+0.11$ & \cellcolor{ULCTgain!33}$+3.24$\\
\faintrule
\multicolumn{12}{l}{\textit{Dravidian} (2)}\\
\quad malayalam & 30.62 & 30.62 & 29.36 & 30.78 & 24.91 & 26.89 & 25.51 & \cellcolor{ULCTgain!45}$+5.71$ & \cellcolor{ULCTgain!45}$+5.71$ & \cellcolor{ULCTgain!41}$+4.44$ & \cellcolor{ULCTgain!45}$+5.87$\\
\quad tamil & 31.29 & 33.00 & 32.22 & 34.09 & 27.18 & 26.96 & 27.89 & \cellcolor{ULCTgain!39}$+4.11$ & \cellcolor{ULCTgain!45}$+5.82$ & \cellcolor{ULCTgain!45}$+5.04$ & \cellcolor{ULCTgain!45}$+6.91$\\
\faintrule
\multicolumn{12}{l}{\textit{Austroasiatic} (1)}\\
\quad vietnamese & 31.04 & 32.58 & 31.67 & 34.80 & 30.96 & 30.09 & 30.07 & $+0.09$ & $+1.62$ & $+0.71$ & \cellcolor{ULCTgain!37}$+3.84$\\
\faintrule
\multicolumn{12}{l}{\textit{Basque} (1)}\\
\quad basque & 29.60 & 31.11 & 31.93 & 32.20 & 31.93 & 29.44 & 30.40 & \cellcolor{ULCTloss!27}$-2.33$ & $-0.82$ & $+0.00$ & $+0.27$\\
\faintrule
\multicolumn{12}{l}{\textit{Japonic} (1)}\\
\quad japanese & 31.02 & 31.73 & 30.78 & 32.22 & 27.29 & 29.49 & 30.38 & \cellcolor{ULCTgain!37}$+3.73$ & \cellcolor{ULCTgain!41}$+4.44$ & \cellcolor{ULCTgain!35}$+3.49$ & \cellcolor{ULCTgain!45}$+4.93$\\
\faintrule
\multicolumn{12}{l}{\textit{Kartvelian} (1)}\\
\quad georgian & 29.47 & 30.42 & 29.58 & 31.00 & 28.16 & 28.27 & 27.69 & \cellcolor{ULCTgain!21}$+1.31$ & \cellcolor{ULCTgain!27}$+2.27$ & $+1.42$ & \cellcolor{ULCTgain!31}$+2.84$\\
\faintrule
\multicolumn{12}{l}{\textit{Koreanic} (1)}\\
\quad korean & 31.80 & 33.42 & 31.78 & 35.13 & 31.36 & 30.38 & 30.96 & $+0.44$ & $+2.07$ & $+0.42$ & \cellcolor{ULCTgain!37}$+3.78$\\
\faintrule
\multicolumn{12}{l}{\textit{Kra-Dai} (1)}\\
\quad thai & 30.11 & 30.87 & 30.36 & 32.04 & 25.76 & 30.56 & 26.96 & \cellcolor{ULCTgain!41}$+4.36$ & \cellcolor{ULCTgain!45}$+5.11$ & \cellcolor{ULCTgain!42}$+4.60$ & \cellcolor{ULCTgain!45}$+6.29$\\
\faintrule
\multicolumn{12}{l}{\textit{Mande} (1)}\\
\quad bambara & 25.16 & 26.71 & 27.44 & 26.96 & 26.47 & 28.33 & 26.58 & $-1.31$ & $+0.24$ & \cellcolor{ULCTgain!18}$+0.98$ & $+0.49$\\
\faintrule
\multicolumn{12}{l}{\textit{Nilotic} (1)}\\
\quad luo & 24.40 & 26.11 & 26.16 & 27.42 & 27.93 & 26.27 & 27.80 & \cellcolor{ULCTloss!35}$-3.53$ & \cellcolor{ULCTloss!24}$-1.82$ & \cellcolor{ULCTloss!24}$-1.78$ & $-0.51$\\
\faintrule
\multicolumn{12}{l}{\textit{Sino-Tibetan} (1)}\\
\quad chinese & 31.33 & 33.96 & 32.56 & 35.69 & 28.96 & 29.96 & 26.20 & \cellcolor{ULCTgain!28}$+2.38$ & \cellcolor{ULCTgain!45}$+5.00$ & \cellcolor{ULCTgain!36}$+3.60$ & \cellcolor{ULCTgain!45}$+6.73$\\
\midrule
\multicolumn{12}{l}{\textbf{XStoryCloze}, \textit{all 9 languages, grouped by family}}\\
\multicolumn{12}{l}{\textit{Indo-European} (4)}\\
\quad english & 58.76 & 57.33 & 56.24 & 56.84 & 57.13 & 56.29 & 56.93 & $+1.63$ & $+0.20$ & $-0.89$ & $-0.29$\\
\quad hindi & 55.94 & 54.11 & 54.75 & 55.10 & 55.34 & 54.75 & 54.71 & $+0.60$ & $-1.23$ & $-0.60$ & $-0.24$\\
\quad russian & 55.59 & 56.33 & 53.90 & 54.33 & 56.68 & 54.97 & 56.84 & $-1.09$ & $-0.34$ & $-2.78$ & $-2.34$\\
\quad spanish & 56.49 & 55.13 & 54.35 & 55.21 & 55.45 & 53.63 & 53.24 & $+1.05$ & $-0.32$ & $-1.10$ & $-0.24$\\
\faintrule
\multicolumn{12}{l}{\textit{Atlantic-Congo} (1)}\\
\quad swahili & 56.73 & 55.01 & 53.98 & 55.02 & 56.31 & 55.31 & 54.72 & $+0.42$ & $-1.30$ & $-2.33$ & $-1.28$\\
\faintrule
\multicolumn{12}{l}{\textit{Afro-Asiatic} (1)}\\
\quad arabic & 52.71 & 52.50 & 51.49 & 51.36 & 50.50 & 51.52 & 51.54 & $+2.21$ & $+2.00$ & $+0.99$ & $+0.86$\\
\faintrule
\multicolumn{12}{l}{\textit{Austronesian} (1)}\\
\quad indonesian & 56.47 & 55.59 & 54.94 & 55.22 & 55.27 & 53.98 & 55.09 & $+1.19$ & $+0.32$ & $-0.33$ & $-0.05$\\
\faintrule
\multicolumn{12}{l}{\textit{Basque} (1)}\\
\quad basque & 55.82 & 54.55 & 53.62 & 53.99 & 53.59 & 53.24 & 55.17 & $+2.22$ & $+0.95$ & $+0.03$ & $+0.40$\\
\faintrule
\multicolumn{12}{l}{\textit{Sino-Tibetan} (1)}\\
\quad chinese & 54.72 & 54.30 & 53.47 & 56.72 & 54.16 & 54.04 & 54.14 & $+0.56$ & $+0.13$ & $-0.69$ & \cellcolor{ULCTgain!29}$+2.55$\\
\midrule
\multicolumn{12}{l}{\textbf{PAWS-X}, \textit{all 7 languages, grouped by family}}\\
\multicolumn{12}{l}{\textit{Indo-European} (4)}\\
\quad english & 90.52 & 90.35 & 90.88 & 90.23 & 89.44 & 89.11 & 90.16 & \cellcolor{ULCTgain!19}$+1.08$ & $+0.91$ & \cellcolor{ULCTgain!22}$+1.44$ & $+0.79$\\
\quad french & 78.65 & 78.58 & 79.70 & 79.02 & 74.53 & 75.31 & 72.09 & \cellcolor{ULCTgain!39}$+4.12$ & \cellcolor{ULCTgain!39}$+4.05$ & \cellcolor{ULCTgain!45}$+5.17$ & \cellcolor{ULCTgain!42}$+4.49$\\
\quad german & 76.99 & 77.89 & 78.89 & 77.85 & 74.89 & 75.41 & 71.40 & \cellcolor{ULCTgain!26}$+2.10$ & \cellcolor{ULCTgain!32}$+3.00$ & \cellcolor{ULCTgain!38}$+4.00$ & \cellcolor{ULCTgain!32}$+2.96$\\
\quad spanish & 77.84 & 77.89 & 78.41 & 78.03 & 76.60 & 75.85 & 72.16 & $+1.24$ & $+1.29$ & \cellcolor{ULCTgain!24}$+1.81$ & $+1.43$\\
\faintrule
\multicolumn{12}{l}{\textit{Japonic} (1)}\\
\quad japanese & 64.25 & 63.88 & 63.27 & 64.33 & 59.44 & 62.51 & 55.83 & \cellcolor{ULCTgain!44}$+4.81$ & \cellcolor{ULCTgain!41}$+4.44$ & \cellcolor{ULCTgain!37}$+3.83$ & \cellcolor{ULCTgain!44}$+4.89$\\
\faintrule
\multicolumn{12}{l}{\textit{Koreanic} (1)}\\
\quad korean & 63.41 & 64.59 & 63.55 & 64.81 & 61.89 & 63.04 & 62.04 & \cellcolor{ULCTgain!22}$+1.52$ & \cellcolor{ULCTgain!30}$+2.70$ & \cellcolor{ULCTgain!23}$+1.66$ & \cellcolor{ULCTgain!31}$+2.92$\\
\faintrule
\multicolumn{12}{l}{\textit{Sino-Tibetan} (1)}\\
\quad chinese & 67.37 & 66.50 & 68.34 & 68.01 & 61.61 & 67.05 & 59.43 & \cellcolor{ULCTgain!45}$+5.76$ & \cellcolor{ULCTgain!44}$+4.89$ & \cellcolor{ULCTgain!45}$+6.73$ & \cellcolor{ULCTgain!45}$+6.40$\\

\bottomrule

\caption{Per-language test accuracy (\%) on every evaluated language, averaged over five fine-tuning seeds. 
$\Delta$ is the paired difference against BPE over the five shared seeds. Green marks a difference in LCT's favour and red against it; only differences that clear a two-sided paired $t$ test at $p<0.05$ across the five seeds are shaded, and saturation scales with the size of the difference up to 5 points. Unshaded cells are within seed-to-seed noise. Languages are grouped by genealogical family, ordered by how many of the 104 pretraining languages each family contributes. Belebele covers the 79 test languages present in the pretraining corpus. 
}\label{tab:fw104-per-language-full}

\end{longtable}


\begin{figure}[h]
    \centering
    \includegraphics[width=0.5\linewidth]{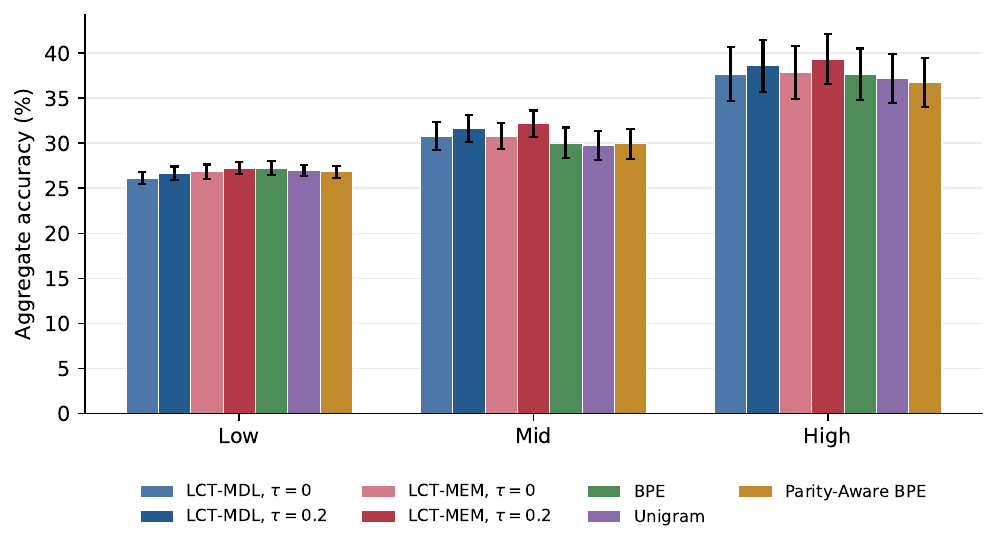}
    \caption{Aggregate downstream accuracy across low-, mid-, and high-resource language tertiles. Scores are averaged across the available XNLI, Belebele, XStoryCloze, and PAWS-X evaluations and five fine-tuning seeds. Error bars indicate 95\% confidence intervals across languages.}
    \label{fig:resource-downstream}
\end{figure}

\end{document}